\documentclass[lettersize,journal]{IEEEtran}
\usepackage{amsmath,amsfonts}
\usepackage{algorithmic}
\usepackage{algorithm}
\usepackage{array}
\usepackage[caption=false,font=normalsize,labelfont=sf,textfont=sf]{subfig}
\usepackage{textcomp}
\usepackage{stfloats}
\usepackage{url}
\usepackage{verbatim}
\usepackage{graphicx}
\usepackage{cite}
\usepackage{booktabs,multirow,adjustbox,diagbox,threeparttable}
\usepackage[colorlinks,linkcolor=black,citecolor=black]{hyperref}
\usepackage{xcolor}
\newcommand{\revised}[1]{{\textcolor{black}{#1}}}
\begin{document}

\title{Predictive Sample Assignment for Semantically Coherent Out-of-Distribution Detection}

\author{Zhimao~Peng,~Enguang~Wang, Xialei~Liu,~\IEEEmembership{Member,~IEEE}, \\ and~Ming-Ming~Cheng,~\IEEEmembership{Senior Member,~IEEE}
\IEEEcompsocitemizethanks{\IEEEcompsocthanksitem Zhimao Peng, Enguang Wang, Xialei Liu, and Ming-Ming Cheng are with Nankai University.
\IEEEcompsocthanksitem Corresponding author: xialei@nankai.edu.cn
\IEEEcompsocthanksitem 
\IEEEcompsocthanksitem 
}}

\markboth{Journal of \LaTeX\ Class Files,~Vol.~14, No.~8, August~2021}%
{Shell \MakeLowercase{\textit{et al.}}: A Sample Article Using IEEEtran.cls for IEEE Journals}


\maketitle

\begin{abstract}
Semantically coherent out-of-distribution detection (SCOOD) is a recently proposed realistic  OOD detection setting: given labeled in-distribution (ID) data and mixed in-distribution and out-of-distribution unlabeled data as the training data, SCOOD aims to enable the trained model to accurately identify OOD samples in the testing data. Current SCOOD methods mainly adopt various clustering-based in-distribution sample filtering (IDF) strategies to select clean ID samples from unlabeled data, and take the remaining samples as auxiliary OOD data, which inevitably introduces a large number of noisy samples in training. To address the above issue, we propose a concise SCOOD framework based on predictive sample assignment (PSA). PSA includes a dual-threshold ternary sample assignment strategy based on the predictive energy score that can significantly improve the purity of the selected ID and OOD sample sets by assigning unconfident unlabeled data to an additional discard sample set, and a concept contrastive representation learning loss to further expand the distance between ID and OOD samples in the representation space to assist ID/OOD discrimination. In addition, we also introduce a retraining strategy to help the model fully fit the selected auxiliary ID/OOD samples. Experiments on two standard SCOOD benchmarks demonstrate that our approach outperforms the state-of-the-art methods by a significant margin. \revised{The code is available at: \href{https://github.com/ZhimaoPeng/PSA}{https://github.com/ZhimaoPeng/PSA}}.
\end{abstract}


\section{Introduction}
\IEEEPARstart{D}{eep} learning models deployed and running in the open world often encounter out-of-distribution (OOD) data whose categories do not overlap with those of in-distribution (ID) data \cite{hendrycks2016baseline,lee2018simple}. The goal of OOD detection is to enable the model to correctly predict ID data and identify OOD data as abnormal samples \cite{bulusu2020anomalous,yang2021semantically}. Although OOD detection has attracted wide attention in the field of machine learning and multimedia due to its key role in security applications such as autonomous driving \cite{geiger2012we,FangQXL24,CenJXJST24}, identity authentication \cite{zhong2019centralized,leng2019survey}, abnormal event detection\cite{zhong2022bidirectional} and fraud detection \cite{choi2018generative}, it is still very challenging because modern neural networks always make overconfident predictions on OOD data \cite{nguyen2015deep}.

\begin{figure}
    \centering
    \includegraphics[width=0.99\columnwidth]{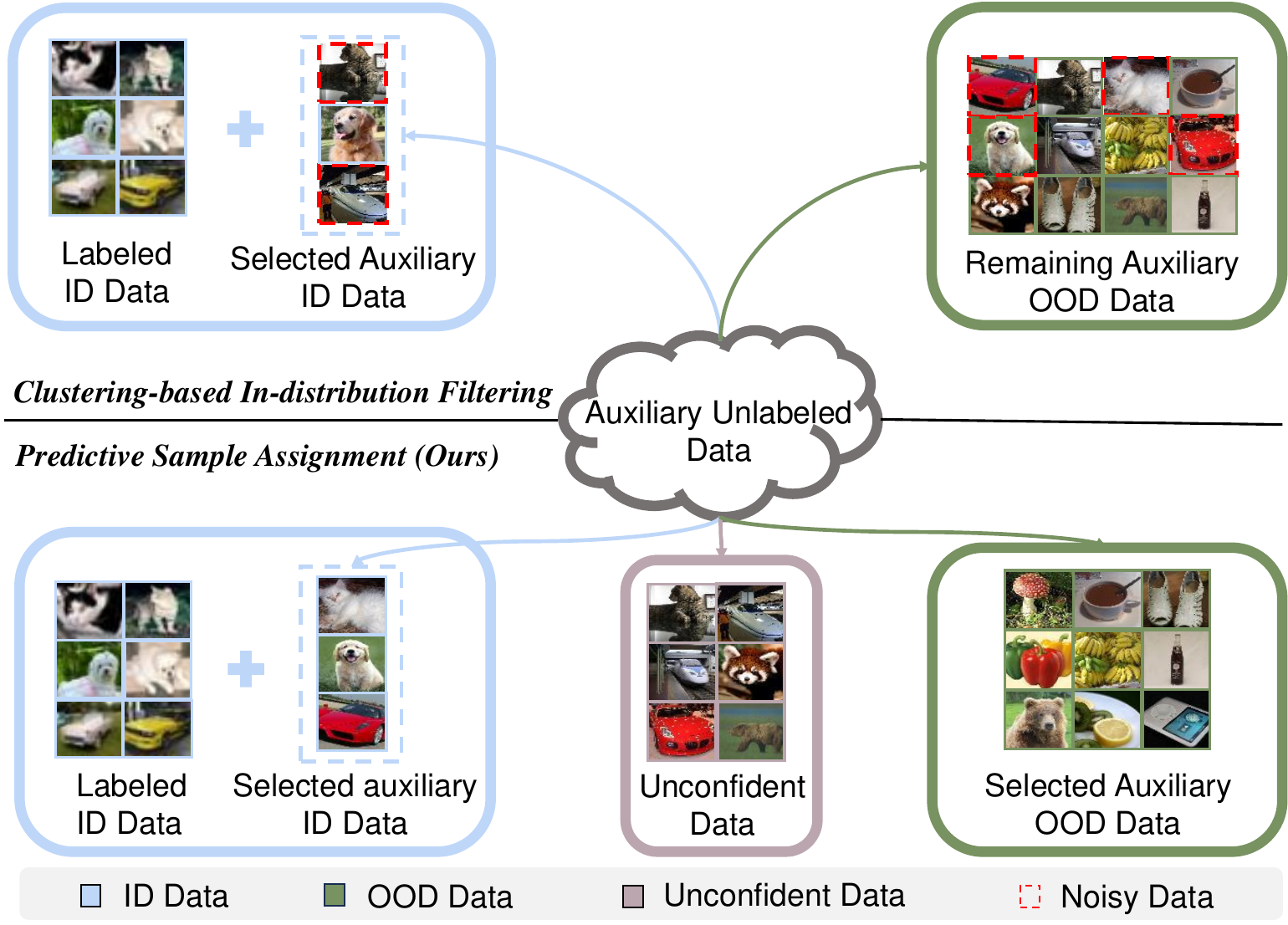}
    \caption{Clustering-based In-distribution Filtering (IDF) vs. Predictive Sample Assignment (PSA). IDF roughly divides unlabeled data into auxiliary ID and OOD samples, which inevitably introduces a large number of noisy samples, and adding them into SCOOD training will lead to sub-optimal performance. In contrast, our PSA excludes training of unconfident samples by assigning them to an additional discard sample set, allowing the model to learn semantically more discriminative representations, thereby improving the SCOOD performance.}
    \label{fig:motivation}
\end{figure}

In order to alleviate the overconfidence problem of the model, outlier exposure \cite{hendrycks2018deep} attempts to introduce auxiliary OOD data (usually another dataset) into the model training, so that the model can explicitly learn to recognize ID and OOD patterns. Although significantly improved OOD detection results have been achieved, it is not easy to collect ideal auxiliary OOD data in reality, because the original auxiliary OOD data (crawled from the Internet) often contains ID data, and the purification cost of these auxiliary OOD data manually is high. The direct use of these unpurified auxiliary OOD data for training will cause the model to overfit these samples, so that the model will excessively focus on the low-level covariate shift between datasets and ignore the high-level semantic information, resulting in poor OOD detection performance when the testing data comes from unseen domains.

To alleviate the above problems, SCOOD benchmarks \cite{yang2021semantically} are proposed to promote the model to perform OOD detection by discriminating semantic differences between ID/OOD samples: the unlabeled auxiliary OOD dataset contains a part of the samples whose categories overlap with the ID dataset. The existing methods \cite{yang2021semantically,lu2023uncertainty} mainly adopt the strategy of clustering-based ID sample filtering (IDF) to select clean ID samples from noisy auxiliary OOD dataset: in each epoch of training, unsupervised clustering ($k$-means \cite{macqueen1967some} or optimal transport \cite{cuturi2013sinkhorn} based on energy score) is carried out on both ID and OOD samples. If samples of an ID category in a sample cluster dominate the cluster, all OOD samples in the cluster are regarded as samples of that ID category and added to the classification training. The remaining OOD samples will be used as the auxiliary OOD data for outlier exposure training. Although promising results have been achieved, the clustering-based IDF method is sub-optimal for filtering ID samples: due to the poor visual representation ability of the model in the early stage of training, using IDF to roughly divide the auxiliary OOD samples into ID and OOD samples in each epoch will make them mixed with a large number of noise samples, which will lead to inferior representation learning of the model and damage the IDF process in the next epoch. This cumulative error will lead to poor OOD detection performance of the model.  

In order to mitigate the introduction of of noisy samples in the training of SCOOD tasks, in this paper, we propose a concise SCOOD framework based on a predictive sample assignment (PSA) strategy (see Fig. \ref{fig:motivation}). Specifically, instead of using the clustering-based IDF strategy to make the binary decision of clean ID/OOD for the auxiliary OOD samples, we perform the dual-threshold ternary assignment based on the predicted energy score of the samples: the auxiliary OOD samples are divided into ID samples, OOD samples and unconfident samples. By excluding unconfident samples from training and employing energy score as the criterion, the model can use less noisy auxiliary ID and OOD samples for model training. In addition, we introduce a concept contrastive representation learning loss (CCL), which enables the model to further expand the distance between ID and OOD samples in the representation space by taking all assigned OOD samples as a single semantic concept, thus promoting the discrimination ability of the model for ID/OOD samples. Finally, to fully fit the selected ID/OOD samples, we propose to use the final selected auxiliary samples and the labeled ID samples for model retraining, which further improves the performance.  

The main contributions of this paper are summarized as follows: \textbf{1)} We analyze the limitations of existing SCOOD methods based on the clustering-based binary In-distribution Filtering (IDF) strategy and propose to use dual-threshold ternary sample assignment to select more reliable auxiliary ID and OOD samples for the SCOOD task.  \textbf{2)} We propose a predictive sample assignment (PSA) framework for the SCOOD task, which includes a threshold sample assignment mechanism based on predicted energy score, a ID/OOD concept contrastive representation learning loss, and an additional sample retraining process, so that the model can fully learn from reliable auxiliary unlabeled samples. \textbf{3)} We conduct extensive experiments on standard SCOOD benchmarks, the
results demonstrate our approach achieves state-of-the-art performance. 

\section{Related works}
\subsection{OOD Detection without Unknown Knowledge}
As a typical binary classification problem, the aim of OOD detection is to determine whether an input sample in the testing stage comes from the in-distribution or not. Many studies focused on the design of post-hoc methods based on the response of the model \cite{hendrycks2016baseline,liang2017enhancing,lee2018simple,liu2020energy,huang2021importance,sun2021react,sun2022dice,wang2022vim,wei2022mitigating,zhu2022boosting,ahn2023line,yu2023block,djurisic2023extremely,sun2024classifier}. The seminal work \cite{hendrycks2016baseline} takes softmax as the scoring function, the insight behind this is that the softmax classification probability of the OOD sample is statistically lower than that of the ID sample. ODIN \cite{liang2017enhancing} expands the difference between ID and OOD data by adding small perturbations to input samples and adjusting the temperature parameter of the softmax function.  Lee \textit{et al.} \cite{lee2018simple} analyzes the output of the model and the response of the middle layer, and designs the scoring function based on the likelihood or density for OOD detection. Gradient information \cite{huang2021importance} and energy score \cite{liu2020energy} have also been shown to be useful as a score function to determine whether a sample is an OOD sample or not. Sun \textit{et al.} \cite{sun2021react} designed a simple rectified activation method to reduce the oversized activation of hidden units, thereby reducing the overconfidence of OOD data. \cite{wei2022mitigating} performed logit normalization by setting a fixed vector norm to the logits during model training, thus alleviating the overconfidence for OOD data.  DICE\cite{sun2022dice} selectively used the most significant model weights to compute the output for OOD detection, thereby reducing the output variance of OOD data and enhancing the separability from ID data.   \cite{sun2024classifier} reduced feature activation of OOD data by masking features that are less important to the ID class, thereby widening the scoring gap between ID and OOD data. Recently, some non-parametric \cite{sun2022out} and representation learning methods \cite{tack2020csi,sehwag2021ssd,huang2021mos,ming2022exploit,sun2022moep} have also been proposed to enhance OOD detection. In addition, \cite{fang2024learnability} studied the learnability of OOD detection in real scenarious. which provided a theoretical basis for the current OOD detection methods.
\subsection{OOD Detection with Outlier Exposure}

The OOD detection methods based on outlier exposure attempt to learn the difference between ID and OOD samples from additional OOD samples. For scenarios where OOD samples are not available, some works \cite{lee2017training,vernekar2019out,du2022vos,tao2023non,JiangZWH23,zheng2023out,du2024dream} have studied the synthesis strategy of OOD training data. VOS \cite{du2022vos} first estimates the class-conditional distribution in the representation space, and the outliers can be sampled from the low-likelihood region of the ID class. Tao \textit{et al.} \cite{tao2023non} proposed a non-parametric outlier synthesis method to avoid making any distribution assumption about the ID embeddings. Jiang \textit{et al.} \cite{JiangZWH23} synthesized high-quality outliers by mixing randomly rotated version of the samples themselves. For scenarios where the auxiliary OOD training samples are provided directly\cite{hendrycks2018deep,yu2019unsupervised,chen2021atom,ming2022poem,wang2023out,wang2024learning,zhu2024diversified}, OE \cite{hendrycks2018deep} encourages the high entropy prediction of OOD samples on the classifier, and MCD \cite{yu2019unsupervised} explicitly enlarges the entropy difference between ID and OOD samples. \cite{chen2021atom,ming2022poem} studied sampling strategies for efficient learning of auxiliary OOD samples. \cite{wang2023out,wang2024learning,zhu2024diversified} used existing auxiliary OOD samples to synthesis more informative outliers to bridge the gap between auxiliary OOD distribution and real OOD distribution. Despite the significant performance gains achieved, however, the auxiliary OOD samples in the current setting are from another dataset that is not the same as the ID samples. As a result, the OOD detection model tends to overfit the covariate shift between datasets, which makes it difficult to guarantee the OOD detection performance of the model when facing the testing samples from the unknown domain. To alleviate this problem, the SCOOD benchmark \cite{yang2021semantically} assumes that the unlabeled auxiliary OOD dataset is mixed with a part of ID samples, so that the model should be able to understand the semantic differences between ID and OOD samples to separate them. For this purpose, UDG \cite{yang2021semantically} and ET-OOD \cite{lu2023uncertainty} are proposed to use the cluster-based ($k$-means, optimal transport) in-distribution sample filtering strategy to select ID samples from unlabeled data. Unlike these methods, which directly divide the unlabeled samples into ID and OOD data, we propose a ternary assignment of samples to make the assigned ID and OOD samples more reliable. In addition, a similar setting to SCOOD has been proposed and modeled as a constrained optimization problem for solving \cite{katz2022training}.

\subsection{Energy Score for OOD Detection}
Recently, energy score \cite{lecun2006tutorial} has been widely used in OOD detection due to its simplicity and direct compatibility with deep neural networks. This score reflects the overall uncertainty of the model for a given input, with a lower energy score indicating more confidence for the prediction and a higher energy score indicating less certainty for the prediction. In \cite{liu2020energy}, the energy score is used as a scoring function directly for OOD detection. Will \textit{et al.} \cite{grathwohl2019your} propose to use the norm of the gradient of energy score as the scoring function. Lin \textit{et al.} \cite{lin2021mood} propose to set up multiple OOD detectors at different depths of the model, and introduce an adjusted energy score as the scoring function to adapt to this multi-level out-of-distribution detection. Wang \textit{et al.} \cite{wang2021can} propose to aggregate the energy scores from multiple labels as the scoring function for OOD detection under the multi-label classification setting. Wu \textit{et al.} \cite{wu2023energybased} designed an efficient OOD discriminator for OOD detection on graph neural network based on energy function. ET-OOD \cite{lu2023uncertainty} propose the optimal transport based on energy score to optimize the cluster assignment of training samples, and separating more ID samples from unlabeled data, thus improving the performance of SCOOD. By contrast, our method proposes to use a dual-threshold ternary sample assignment framework based on predictive energy scores to select reliable ID and OOD samples from unlabeled data, making the auxiliary samples used for SCOOD training less noisy. 

\section{Preliminaries}
\subsection{Problem Formulation}
Given the training data $\mathcal{D} = \mathcal{D}_L \cup \mathcal{D}_U$, where $\mathcal{D}_L=\left\{\mathbf{x}_i, {y}_i\right\}_{i=1}^n$ is a labeled dataset and $\mathcal{D}_U=\left\{\mathbf{x}_i\right\}_{i=1}^m$ is an auxiliary unlabeled dataset. In the previous OOD setting, all samples in $\mathcal{D}_L$ come from in-distribution $\mathcal{I}$ and all samples in $\mathcal{D}_U$ come from out-of-distribution $\mathcal{O}$. We define the category spaces of $\mathcal{I}$ and $\mathcal{O}$ as $\mathcal{C}^I$ and $\mathcal{C}^O$ respectively and there is no overlap between $\mathcal{C}^I$ and $\mathcal{C}^O$. But in the more realistic SCOOD setting, unlabeled dataset $\mathcal{D}_U= \mathcal{D}_U^I \cup \mathcal{D}_U^O$, where $\mathcal{D}_U^I \subset \mathcal{I}$ and $\mathcal{D}_U^O \subset \mathcal{O}$. The testing data  $\mathcal{T}$ comprises $\mathcal{T}^I$ and $\mathcal{T}^O$, where $\mathcal{T}^I \subset \mathcal{I}$ and $\mathcal{T}^O \subset \mathcal{O}$. It is worth noting that the category spaces of $\mathcal{D}_U^O$ and $\mathcal{T}^O$ could be non-overlapping. The goal of SCOOD is not only to reject samples in $\mathcal{T}^O$ but to correctly classify samples in $\mathcal{T}^I$ by employing a model trained with all training data $\mathcal{D}$.

\subsection{In-distribution Filtering for SCOOD}
Current SCOOD methods \cite{yang2021semantically,lu2023uncertainty} adopt an In-distribution Filtering (IDF) strategy to select ID samples from the unlabeled dataset. Specifically, at epoch $t$, IDF first groups all the training samples $\mathcal{D}$ based on the clustering method (\textit{i.e., k-means, optimal transport}), the resulting $k$-th sample cluster can be represented as $\mathcal{D}_k^{(t)}$. The proportion of samples with class $c \subset \mathcal{C}^I$ in cluster $k$ can be calculated as:

\begin{equation}
rate_{k, c}^{(t)}=\frac{\left| \mathcal{D}_{k, y}^{(t)}=\left\{x_i \mid index=k, y_i=c\right\}\right|}{\left|\mathcal{D}_k^{(t)}\right|}
\end{equation}

if ${rate}_{k, c}^{(t)}$ exceeds the pre-defined threshold $\tau$, all unlabeled samples in cluster $k$ will be merged into the labeled dataset with label $c$. Finally, the updated labeled dataset  $\mathcal{D}_L^{(t)}$ can be defined as:

\begin{equation}
\mathcal{D}_L^{(t)}=\mathcal{D}_L \cup\left\{x_i \mid x_i \in \mathcal{D}_k^{(t)}, \text { rate }_{k, c}^{(t)}>\tau\right\}
\end{equation}

In the meanwhile, all remaining unlabeled samples will be updated as $\mathcal{D}_U^{(t)}$. During the SCOOD training, outlier exposure (OE) is used to boost the detection capability by learning from the labeled dataset $D_L^{(t)}$ and surrogate unlabeled dataset $\mathcal{D}_U^{(t)}$, with the associated learning objective as follows:

\begin{equation}
\mathcal{L}=\mathcal{L}_{\text {CE}} +\gamma \mathcal{L}_{\text {OE}}+\lambda \mathcal{L}_{\text {Aux}}
\end{equation}

\begin{equation}
\mathcal{L}_{\text {CE}} = \frac{1}{\left|\mathcal{D}_L^{(t)}\right|} \sum_{x_i \in \mathcal{D}_L^{(t)}} \ell\left(\boldsymbol{y}_i, \boldsymbol{p(x_i)}\right)
\end{equation}

\begin{equation}
\mathcal{L}_{\text {OE}} = 
 \frac{1}{\left|\mathcal{D}_U^{(t)}\right|} \frac{1}{\left|\mathcal{C}^I\right|} \sum_{x_i \in \mathcal{D}_U^{(t)}} \sum_{{c}_i \in \mathcal{C}^I}  \ell\left(\boldsymbol{c}_i, \boldsymbol{p(x_i)}\right)
\end{equation}

\noindent where, $\gamma$ and $\lambda$ are trade-off parameters, $\boldsymbol{p(x_i)}$ is the softmax normalized 
 classification probability of $x_i$. $\ell$ is cross-entropy loss. $\mathcal{L}_{\text {OE}}$ is a regularization loss term which makes model learn from $\mathcal{D}_U^{(t)}$ with low confident predictions. $\mathcal{L}_{\text {Aux}}$ is an auxiliary representation learning loss that varies across different methods.

\subsection{Energy Score}
Energy score \cite{liu2020energy} is a simple and efficient scoring function in the OOD detection task, which can be formulated as:

\begin{equation}
E(x_i)=-T \cdot \log \sum_{{c}_i \in \mathcal{C}^I} e^{l\left(c_i \mid x_i\right) / T}
\end{equation}

\noindent where $T$ is the temperature parameter, $l\left(c_i \mid x_i\right)$ indicates the logit of $x_i$ on class $c_i$. From the above equation, it can be seen that the energy score is directly related to the probability of the input sample, that is, an in-distribution sample will usually have a lower energy score.

\begin{figure*}[t]
    \centering
    \includegraphics[width=0.95\textwidth]{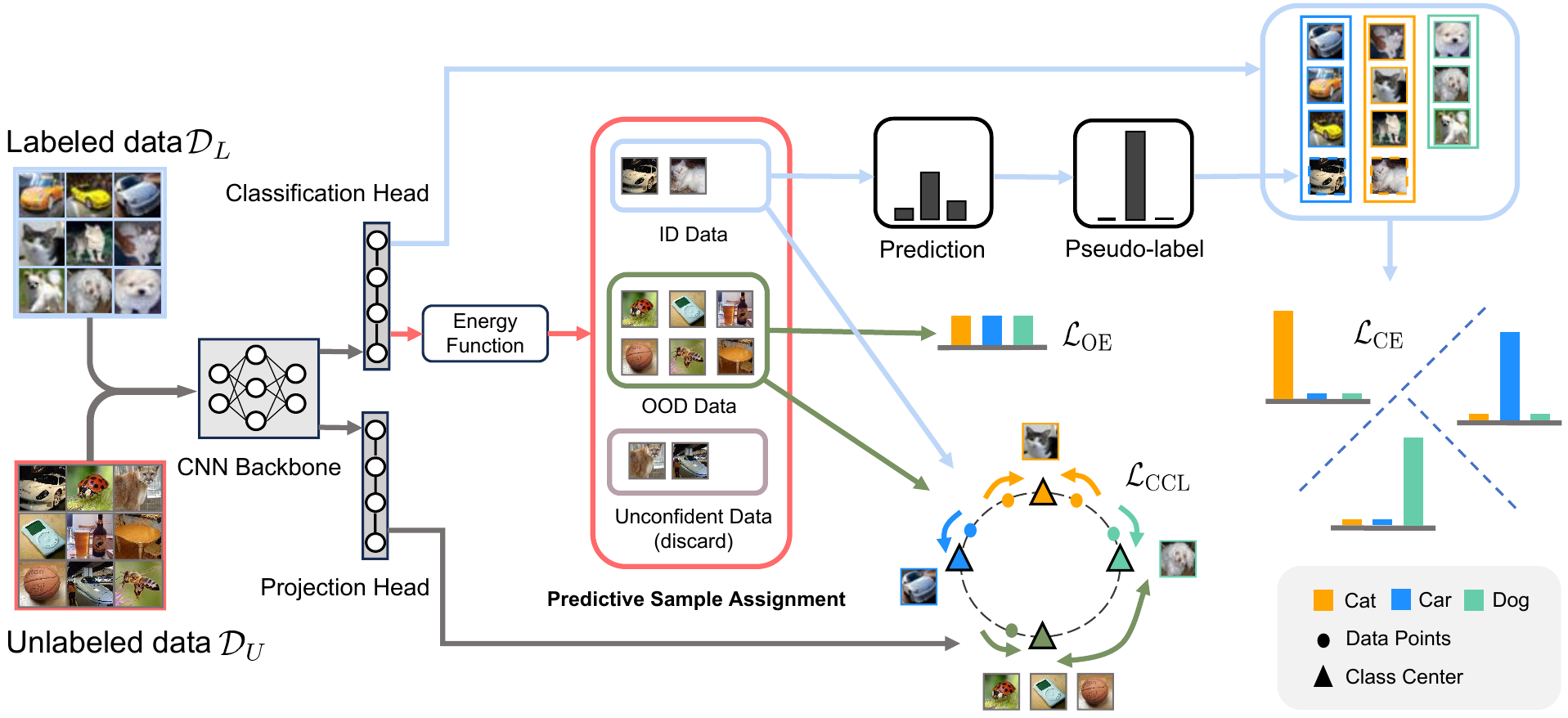}
    \vspace{-.1in}
    \caption{ Illustration of the proposed predictive sample assignment (PSA) framework for SCOOD. The full CNN model consists of a backbone $f$, a classification head $l$, and a projection head $g$. The energy score of unlabeled data can be calculated from the logits output of $l$, and then a ternary sample assignment based on dual-threshold is performed by PSA. During the model training, classification loss $\mathcal{L}_{\text {CE}}$ is calculated based on the logits output of the selected ID data and the labeled data on the classification head $l$. Outlier exposure loss $\mathcal{L}_{\text {OE}}$ is calculated based on the logits output of the selected OOD data on the classification head $l$. ID/OOD concept contrastive loss $\mathcal{L}_{\text {CCL}}$ is calculated based on the representation output of the selected ID/OOD data and the labeled data on the projection head $g$. 
    }
    \label{fig:method}
\end{figure*}

\section{Our Method}
In this section, we will elaborate on our proposed predictive sample assignment (PSA) framework for SCOOD. Following the basic framework of SCOOD, the model learns to identify ID samples while learning to output low confidence predictions for OOD samples. Different from previous methods that only filter clean ID samples, we introduce predictive sample assignment in Section \ref{subsec:psa_framework} to simultaneously select clean ID and OOD samples from the unlabeled dataset. In order to further widen the distance between different semantic concepts in the representation space, we introduce a concept contrastive learning loss  (CCL), which is introduced in Section \ref{subsec:ccl_loss}. Finally, to make full use of the auxiliary unlabeled data and further improve performance, we retrain the model with selected confident ID/OOD data and labeled training data, this procedure is elaborated in Section \ref{subsec:retrain_procedure}. Fig. \ref{fig:method} illustrates the proposed PSA framework.

\subsection{Predictive Sample Assignment}
\label{subsec:psa_framework}
In order to alleviate the biased ID/OOD discriminative representation learning caused by the lack of uncertain sample estimation in current SCOOD methods and simplify the complex clustering-based IDF operations, we propose a concise predictive sample assignment (PSA) strategy to select clean ID samples, unconfident samples and clean OOD samples from unlabeled dataset simultaneously. Therefore, the model can learn from reliable ID and OOD samples, thereby improving the OOD detection capability. Specifically, at epoch $t$, given the negative value of output energy score $s_i$ of an unlabeled sample $x_i$, we can assign unlabeled samples according to a dual-threshold criterion:

\begin{equation}
f(x_i) = \begin{cases} 
\text{ID Sample}, & \text{if } s_i > \delta_{id} \\
\text{OOD Sample}, & \text{if } s_i < \delta_{ood} \\
\text{Unconfident Sample}, & \text{otherwise}
\end{cases}
\label{eq:psa}
\end{equation}

\noindent where $\delta_{id}$ and $\delta_{ood}$ are pre-defined thresholds. However, when choosing the energy score as the output score, it is difficult to select appropriate $\delta_{id}$ and $\delta_{ood}$ in advance because their values are unbounded. To solve this problem, we propose to use the quantile of the negative value of network output energy scores of all labeled data in the $T_{warm}$-th epoch as the $\delta_{id}$ and $\delta_{ood}$ :

\begin{align}
\delta_{id} = Q(S; q_{id}) \label{eq:threshold_1}\\
\delta_{ood} = Q(S; q_{ood}) \label{eq:threshold_2}
\end{align}

\noindent where $S=\left\{s_1, s_2, \ldots, s_n\right\}$, $Q(S, q)$ represents the $q$-th quantile computed from the set $S$ and $0<q<1$. $T_{warm}$ indicates the number of warm-up epochs in model training. 

Finally, the updated labeled dataset $\mathcal{D}_L^{(t)}$ and unlabeled dataset $\mathcal{D}_U^{(t)}$ can be defined as:
\begin{align}
\mathcal{D}_L^{sel} = \left\{x_i \mid x_i \in \mathcal{D}_U, s_i > \delta_{id}\right\} \label{eq:update_1}\\
\mathcal{D}_L^{(t)}=\mathcal{D}_L \cup \mathcal{D}_L^{sel} \label{eq:update_2}\\
\mathcal{D}_U^{(t)}= \left\{x_i \mid x_i \in \mathcal{D}_U, s_i < \delta_{ood}\right\} \label{eq:update_3}
\end{align}

For all selected ID samples $D_L^{sel}$, the process of pseudo-labeling can be formulated as:

\begin{equation}
y_i=\arg \max \left(p\left(x_i\right)\right), \quad x_i \in \mathcal{D}_L^{sel}
\end{equation}

\subsection{ID/OOD Concept Contrastive Representation Learning}
In order to further improve the OOD detection performance of the model, it is helpful to introduce an additional auxiliary representation learning loss to obtain more discriminative feature representation. Previous methods use deep clustering \cite{caron2018deep} or infoNCE \cite{oord2018representation} loss for all training data as auxiliary representation learning loss. In this paper, we propose a concept contrastive learning (CCL) loss to help the model learn more discriminative feature representation. Specifically, given the training data $\mathcal{D} = \mathcal{D}_L^{(t)} \cup \mathcal{D}_U^{(t)}$, we assume $\boldsymbol{z}_i=g\left(f\left(\boldsymbol{x}_i\right)\right)$ is the $\ell_2$-normalized feature embedding of ${x}_i$ in a mini-batch $B$, $f, g$ indicate the backbone network and projection head. The purpose of concept contrastive learning loss in the SCOOD setting is to make samples with the same semantic concepts close while samples with different semantic concepts are far away in the representation space, which can be written as:

\begin{equation}
\mathcal{L}_{\text {CCL}} =\frac{1}{\left|B\right|} \sum_{i \in B} \frac{1}{\left|\mathcal{P}_i\right|} \sum_{q \in \mathcal{P}_i}-\log \frac{\exp \left(\boldsymbol{z}_i^{\top} \boldsymbol{z}_q / \tau_s\right)}{\sum_i^{i \neq n} \exp \left(\boldsymbol{z}_i^{\top} \boldsymbol{z}_n / \tau_s\right)}
\end{equation}

\noindent where $\mathcal{P}_i$ indexes all other images in the mini-batch $B$ that have the same semantic concept as $x_i$, $\tau_s$ is a temperature parameter. We consider the ID samples with the same class label to have the same semantic concept, and treat all selected OOD samples as a single semantic concept. Finally, the overall loss can be written as:

\begin{equation}
\mathcal{L}_{\text {ALL}}=\mathcal{L}_{\text {CE}} +\gamma \mathcal{L}_{\text {OE}}+\lambda \mathcal{L}_{\text {CCL}}
\end{equation}

\label{subsec:ccl_loss}
\subsection{Retraining with Selected Confident Samples}
Although PSA selects auxiliary ID/OOD samples in each epoch and adds them to the training of the next epoch, the learning rate of the model is relatively large in the early stage of training, which makes the model unable to fully fit the training data, resulting in less auxiliary ID samples selected by PSA. Even if PSA selects more reliable auxiliary ID samples in the later stage of training, the model cannot make full use of these data for training, resulting in sub-optimal performance. In order to further improve the model performance, we propose to retrain the model with the selected confident ID/OOD samples by PSA in the last epoch and labeled training data. Specifically, we define the updated labeled dataset and unlabeled dataset by PSA in the last epoch as $\mathcal{D}_L^{(last)}$ and $\mathcal{D}_U^{(last)}$, and the corresponding retraining loss functions of PSA are updated as follows:

\begin{equation}
\mathcal{L}_{\text {CE}}^{re} = \frac{1}{\left|\mathcal{D}_L^{(last)}\right|} \sum_{x_i \in \mathcal{D}_L^{(last)}} \ell\left(\boldsymbol{y}_i, \boldsymbol{p(x_i)}\right)
\end{equation}

\begin{equation}
\mathcal{L}_{\text {OE}}^{re} = 
 \frac{1}{\left|\mathcal{D}_U^{(last)}\right|} \frac{1}{\left|\mathcal{C}^I\right|} \sum_{x_i \in \mathcal{D}_U^{(last)}} \sum_{{c}_i \in \mathcal{C}^I}  \ell\left(\boldsymbol{c}_i, \boldsymbol{p(x_i)}\right)
\end{equation}

\begin{equation}
\mathcal{L}_{\text {PSA}}=\mathcal{L}_{\text {CE}}^{re} +\gamma \mathcal{L}_{\text {OE}}^{re}+\lambda \mathcal{L}_{\text {CCL}}
\end{equation}

\noindent where the mini-batch $B$ in $L_{\text {PSA}}$ is sampled from training data $\mathcal{D} =  \mathcal{D}_L^{(last)} \cup \mathcal{D}_U^{(last)}$.
Algorithm~\ref{alg:algorithm1} lists the pseudo-code of PSA. 

\begin{algorithm}[t]
    \small
	\renewcommand{\algorithmicrequire}{\textbf{Input:}}
	\renewcommand{\algorithmicensure}{\textbf{Output:}}
	\caption{Predictive Sample Assignment for Semantically Coherent Out-of-Distribution Detection.}
	\label{alg:algorithm1}
	\begin{algorithmic}[1]
		\REQUIRE Labeled dataset $\mathcal{D}_L=\left\{\mathbf{x}_i, {y}_i\right\}_{i=1}^n$ , auxiliary unlabeled dataset $\mathcal{D}_U=\left\{\mathbf{x}_i\right\}_{i=1}^m$, loss weight $\gamma,\lambda$, warm-up epochs $T_{warm}$, max epochs $T_{max}$.
		\ENSURE CNN backbone $f$, Classification head $l$.
		\FOR {$t=1,2,...,T_{max}$}
		\IF{$t\leq T_{warm}$}{
			\STATE Train $f$ and $l$ on $\mathcal{D}_L$ with $\mathcal{L}_{\mathrm{CE}}$ and $\mathcal{L}_{\mathrm{CCL}}$.}
		\ELSE
            \STATE Calculate sample assignment thresholds $\delta_{id}$ and $\delta_{ood}$ by eq. \ref{eq:threshold_1} and eq. \ref{eq:threshold_2}. 
		\STATE Selecting auxiliary ID data, unconfident data and auxiliary OOD data by eq. \ref{eq:psa}.
		\STATE Obtain updated labeled dataset $\mathcal{D}_L^{(t)}$ and unlabeled dataset $\mathcal{D}_U^{(t)}$ by eq. \ref{eq:update_1}, eq. \ref{eq:update_2} and eq. \ref{eq:update_3}.
		\STATE Train $f$ and $l$ on $\mathcal{D}_L^{(t)}$ and $\mathcal{D}_U^{(t)}$ with $\mathcal{L}_{\text {ALL}}$.
		\ENDIF
		\ENDFOR
            \STATE Obtain updated labeled dataset and unlabeled dataset by PSA in the last epoch $\mathcal{D}_L^{(\text {last })}$ and $\mathcal{D}_U^{(\text {last })}$.
            \FOR {$t=1,2,...,T_{max}$}
            \STATE Train $f$ and $l$ on $\mathcal{D}_L^{(\text {last })}$ and $\mathcal{D}_U^{(\text {last })}$ with $\mathcal{L}_{\text {PSA}}$.
            \ENDFOR
		\STATE \textbf{return} CNN backbone $f$, Classification head $l$.
	\end{algorithmic}
\end{algorithm}

\label{subsec:retrain_procedure}
\section{Experiments}
\subsection{Experimental Setup}

\textbf{Benchmarks.} Following previous works \cite{yang2021semantically,lu2023uncertainty}, we evaluate our method on two SCOOD benchmarks: CIFAR-10 benchmark, which takes CIFAR-10 as the labeled set $\mathcal{D}_L$ and CIFAR-100 benchmark, which takes CIFAR-100 as labeled set $\mathcal{D}_L$. Tiny-ImageNet is used as an auxiliary unlabeled dataset $\mathcal{D}_U$ for all SCOOD benchmarks. During the testing stage, CIFAR-10/CIFAR-100 is used as one of the OOD testing datasets of the CIFAR-100/CIFAR-10 benchmark respectively. In addition to this, there are five other datasets that serve as testing datasets for all SCOOD benchmarks, including Texture \cite{cimpoi2014describing}, SVHN \cite{netzer2011reading}, Tiny-ImageNet \cite{le2015tiny}, LSUN \cite{yu2015lsun}, and Places365 \cite{zhou2017places}, since they contain both ID and OOD samples, they need to be re-split into $\mathcal{T}^O$ and $\mathcal{T}^I$ according to their true semantic categories for SCOOD benchmarks.

\textbf{Evaluation Metrics.} Following \cite{lu2023uncertainty}, we evaluate the overall performance of PSA with the following six metrics. 
\textbf{FPR95} represents the false positive rate of OOD testing data when the true positive rate of ID testing data is \%95. 
\textbf{AUROC} represents the area under the receiver operating characteristic curve, which takes into account all possible classification thresholds and provides a comprehensive assessment of OOD detection performance. 
\textbf{AUPR-In/Out} represents the area under the precision-recall curve, and when calculating AUPR-In/Out, the ID/OOD testing OOD data are considered as the positive category, respectively. \textbf{CCR@FPR}$n$  represents Correct Classification Rate at a given False Positive Rate of n, which comprehensively evaluates the ID classification and OOD detection capabilities. \textbf{ACC} represents the classification accuracy on all the ID testing data. 

\begin{table*}[btp!]
\renewcommand{\arraystretch}{1.}
\renewcommand{\tabcolsep}{8.pt}
\centering
\small
\caption{\textbf{Comparison with the previous SOTA methods on the two SCOOD benchmarks.} We report the averaged results on all 6 OOD datasets. $\uparrow$/$\downarrow$ indicates higher/lower value is preferred. $\dagger$ indicates the result of reimplementation using their code. The best results are shown in bold font.}
\label{T:Results}
\begin{tabular}{cc|ccc|cccc}
\toprule

\multirow{2}{*}{Benchmarks} 
& \multirow{2}{*}{Method} 
& \multirow{2}{*}{FPR95~$\downarrow$} 
& \multirow{2}{*}{AUROC~$\uparrow$} 
& \multirow{2}{*}{AUPR-In/Out~$\uparrow$}
& \multicolumn{4}{c}{CCR@FPR~$\uparrow$}  \\ 
\cmidrule(lr){6-9}
& && && $10^{-4}$  & $10^{-3}$  & $10^{-2}$ & $10^{-1}$                      \\ \midrule
\multirow{6}{*}{\begin{tabular}[c]{@{}c@{}}CIFAR-10\\ Benchmark\end{tabular}}
& ODIN~\cite{liang2017enhancing}         &  52.00 & 82.00 & 73.13~/~85.12 & 0.36 & 1.29& 	6.92    & 39.37 \\ 
& EBO~\cite{liu2020energy}      &  50.03 &	83.83 &	77.15~/~85.11 & 0.49	& 1.93	&9.12	&46.48 \\
&\revised{ReAct~\cite{sun2021react}}      &  \revised{56.46} &	\revised{90.12} & \revised{88.31~/~88.30} & \revised{0.42} &\revised{3.30} & \revised{31.38} & \revised{78.04} \\
&\revised{LogitNorm~\cite{wei2022mitigating}}      &  \revised{46.57} & \revised{92.05} &	\revised{91.97~/~89.71} & \revised{7.78} & \revised{21.53}	& \revised{46.71} & \revised{80.53} \\
& OE~\cite{hendrycks2018deep}  & 50.53	& 88.93	& 87.55~/~87.83	& 13.41	& 20.25	& 33.91	& 68.20 \\
& MCD~\cite{yu2019unsupervised}           &73.02&	83.89&	83.39~/~80.53&	5.41&	12.3&	28.02&	62.02  \\
& UDG~\cite{yang2021semantically}   & 36.22 & 93.78 & 93.61~/~92.61 & 13.87  & 34.48 & 59.97 & 82.14 \\
& ET-OOD (MSP)$\dagger$ ~\cite{lu2023uncertainty}     & 22.79 & 96.27 & 96.26~/~95.53  &  \textbf{37.41} & 56.59 & 73.35 & 87.42    \\ 
& PSA (Ours)     & \textbf{13.06} & \textbf{97.52} & \textbf{97.47}~/~\textbf{96.95}  &  36.70 & \textbf{61.67} & \textbf{79.33} & \textbf{90.60}         \\ 
\midrule
\multirow{6}{*}{\begin{tabular}[c]{@{}c@{}}CIFAR-100\\ Benchmark\end{tabular}}
& ODIN~\cite{liang2017enhancing}        &81.89	&77.98	&78.54~/~72.56	&1.84	&5.65	& 17.77	& 46.73 \\
& EBO~\cite{liu2020energy}    &81.66	&79.31	&80.54~/~72.82	&2.43	&7.26	& 21.41	& 49.39 \\
&\revised{ReAct~\cite{sun2021react}}      &  \revised{77.95} &	\revised{79.80} & \revised{80.58~/~74.12} & \revised{1.21} &\revised{6.90} & \revised{23.51} & \revised{51.24} \\
&\revised{LogitNorm~\cite{wei2022mitigating}}      &  \revised{82.58} & \revised{76.47} &	\revised{77.39 ~/~70.24} & \revised{1.93} & \revised{4.89}	& \revised{15.57} & \revised{44.97} \\
& OE~\cite{hendrycks2018deep} &80.06	&78.46	&80.22~/~71.83	&2.74	&8.37	& 22.18	& 46.75	\\
& MCD~\cite{yu2019unsupervised}          & 85.14	&74.82	&75.93~/~69.14  &1.06   & 4.60  & 16.73 & 41.83 \\
& UDG~\cite{yang2021semantically}   & 75.45 & 79.63  & 80.69~/~74.10  &  3.85  & 8.66 &  20.57 &  44.47          \\ 
& ET-OOD (MSP)$\dagger$ ~\cite{lu2023uncertainty}   & 76.90 & 81.96  & 83.54~/~75.63  &  8.36  & \textbf{18.87} &  \textbf{32.71} &  51.65           \\ 
& PSA (Ours)     & \textbf{60.37} & \textbf{84.53} & \textbf{85.43}~/~\textbf{80.20}  & \textbf{9.52} & 17.88 & 32.41 & \textbf{54.45}        \\ 
\bottomrule
\end{tabular}
\end{table*}

\textbf{Implementation Details.} Following previous works \cite{yang2021semantically,lu2023uncertainty}, a standard ResNet-18 is used as the backbone network for all the experiments. In both predictive sample assignment stage and retraining stage, the network is trained for 200 epochs. We use an SGD optimizer with a momentum value of 0.9 and a weight decay decay of 0.0005. We set the initial learning rate to 0.1 and use a cosine learning rate annealing strategy with warm-up, and we set the warm-up epoch $T_{warm}$ to 30. For the data loader of $\mathcal{D}_L$ and $\mathcal{D}_U$, the batch size is 64 and 128, respectively. Following \cite{yang2021semantically,lu2023uncertainty}, we set $\gamma$ to 0.5 for all experiments. For $\lambda$, we set it to 0.1 for all experiments.  Following the literature \cite{khosla2020supervised,ming2022exploit}, the temperature parameter $\tau_s$ is set to 0.1 and the dimension of projection head $g$ is set to 128. For CIFAR10 benchmark, we set $q_{id}$ to 0.9 and $q_{ood}$ to 0.3. For CIFAR100 benchmark, we set $q_{id}$ to 0.9 and $q_{ood}$ to 0.1.

\subsection{Results on SCOOD Benchmarks}
We compare the proposed PSA with previous state-of-the-art OOD detection methods and the comparison results are shown in Table \ref{T:Results}, where ODIN \cite{liang2017enhancing}, EBO \cite{liu2020energy} \revised{and ReAct \cite{sun2021react} are post-hoc OOD methods without auxiliary unlabeled dataset, LogitNorm \cite{wei2022mitigating} is a representation learning method without auxiliary unlabeled dataset,} while OE \cite{hendrycks2018deep}, 
 MCD \cite{yu2019unsupervised}, UDG \cite{yang2021semantically}  and ET-OOD \cite{lu2023uncertainty} are OOD methods based on outlier exposure.  Following previous works \cite{yang2021semantically,lu2023uncertainty}, we only report the average results on all 6 OOD datasets on the two SCOOD benchmarks. In particular, for a fair comparison, the results of UDG, ET-OOD and PSA presented in Table~\ref{T:Results} are all using MSP as scoring function. Since T-energy is introduced as a new scoring function by ET-OOD, the comparison results based on T-energy are reported in Table ~\ref{T:metrics}. It can be seen that PSA achieves better results on most metrics of two benchmarks. Especially on FPR95 and AUROC, two core metrics for OOD detection, PSA consistency achieves the best results. Compared to ET-OOD, a method also based on energy score, our PSA achieves better results, which well demonstrates the effectiveness of ternary sample assignment by introducing unconfident sample sets.

\subsection{Ablation Study}

\textbf{Impact of each component.}
To study the Influence of each component in our method, we use CIFAR10-benchmark for evaluation and report the results in Table ~\ref{T:ablation}. EXP$\#1$ is the baseline method which trains the model only on CIFAR-10 with a standard cross-entropy (CE) loss. EXP$\#2$ means that the retraining process is not performed. ``w/o ternary assign" in EXP$\#3$ means that no additional discard sample set is introduced when selecting auxiliary ID/OOD samples. EXP$\#4$ \mbox{--} $\#6$ retrain the model with different loss functions of PSA. By comparing EXP$\#1$ and EXP$\#6$, it can be seen that when the model uses an additional unlabeled dataset $\mathcal{D}_U$ containing ID samples and uses PSA to assign reliable ID/OOD samples from $\mathcal{D}_U$ for the model training, the results of OOD detection have been significant improved, which well proves that our method can effectively use the auxiliary unlabeled dataset to make the model pay attention to the high-level semantic shift between ID and OOD. By comparing EXP$\#2$ and EXP$\#6$, it can be seen that the results of all OOD detection metrics are improved, which well confirm that the model can fully fit reliable auxiliary samples selected by PSA by retraining the model. The comparison between EXP$\#3$ and EXP$\#6$ shows that not introducing an additional discard sample set will lead to inferior results because the model will inevitably fit more noisy samples.  By comparing EXP$\#5$ and EXP$\#6$, it can be seen that the FPR95 and AUROC can be further improved by introducing the ID/OOD concept contrastive learning loss, which confirms that expanding the distance between ID and OOD samples in the representation space can help improve OOD detection performance.

\begin{table}[t]
\centering
\renewcommand{\arraystretch}{1.}
\renewcommand{\tabcolsep}{1.pt}
    \small
    \caption{\textbf{The impact of each component.} EXP$\#1$ uses a model trained on CIFAR-10 dataset with standard cross-entropy (CE) loss. EXP$\#2$ means that the retraining process is not performed. ``w/o ternary assign" in EXP$\#3$ means that no additional discard sample set is introduced when selecting auxiliary ID/OOD samples, which is similar to the IDF strategy. EXP$\#4$ \mbox{--} $\#6$ retrain the model with different loss functions using samples selected by PSA.  }
    \label{T:ablation}
    \resizebox{\columnwidth}{!}{%
    \centering
    \begin{tabular}{c@{\hskip 8pt}l|@{\hskip 1pt}c@{\hskip 1pt}c@{\hskip 1pt}c@{\hskip 1pt}@{\hskip 1pt}c}
    	\toprule
    	$\mathcal{D}$ & Components & \footnotesize{FPR95~$\downarrow$} & \footnotesize{AUROC~$\uparrow$} & \footnotesize{AUPR-IN/OUT~$\uparrow$} & \footnotesize{ACC~$\uparrow$} \\
    	\midrule
    	{\footnotesize{CIFAR-10}}     
    	& 1: CE loss  & 46.03 & 92.89 & 93.17~/~91.29 & 94.17  \\
    	\midrule
    	\multirow{5}{*}{\begin{tabular}[c]{@{}c@{}}\footnotesize{CIFAR-10}\\\footnotesize{+}\\\footnotesize{TIN}\end{tabular}}
        & 2: PSA w/o retraining  & 19.00 & 96.74 & 96.48~/~96.45  & 94.42  \\
        & 3: PSA w/o ternary assign.   & 20.49  & 96.87 & 96.97~/~95.89 & 94.13   \\
        & 4: PSA w/o  $\mathcal{L}_{\text {OE}}^{re}$   &  29.01  & 95.67 & 95.73~/~94.77 & 94.42   \\
        & 5: PSA w/o $\mathcal{L}_{\text {rep}}^{re}$  & 14.67   & 97.32 & 97.20~/~96.87 & \textbf{94.51}   \\
        & 6: PSA   & \textbf{13.06} & \textbf{97.52} & \textbf{97.47~/~96.95} & 94.45  \\
    	\bottomrule
    \end{tabular}
    }
\end{table}

\textbf{Impact of retraining strategies.} We apply the retraining strategy to UDG and ET-OOD and name them UDG+ and ET-OOD+, and their results are reported in Table ~\ref{T:Retraining}, we can find that after retraining, most OOD detection results are decreased, this is because they use the IDF strategy to roughly divide the unlabeled auxiliary dataset into ID and OOD samples, resulting in a lot of noise mixed with these samples, making the retraining strategy can not bring performance gain. In contrast, our PSA excludes the unconfident samples from training by using the ternary based sample assignment strategy, making the selected auxiliary ID and OOD samples more reliable, so that the retraining of these samples could help improve the OOD detection performance of the model.

\begin{table}[t]
\centering
\renewcommand{\arraystretch}{1.}
\renewcommand{\tabcolsep}{4.pt}
\small
    \caption{\textbf{The impact of retraining strategy, UDG+ and ET-OOD+ represent retraining versions of UDG and ET-OOD, respectively.} }
    \label{T:Retraining}
    \resizebox{\columnwidth}{!}{%
    \begin{tabular}{cc|cccc}
    	\toprule
    	 $\mathcal{D}$ & Method & FPR95~$\downarrow$ & \footnotesize{AUROC~$\uparrow$} & \footnotesize{AUPR-IN/OUT~$\uparrow$} & \footnotesize{ACC~$\uparrow$} \\
    	\midrule
     \multirow{6}{*}{\rotatebox{90}{CIFAR-10+TIN}} 
    	& UDG                   &  36.22 & 93.78 & 92.61~/~92.94 & 92.28 \\
        & UDG+                   & 44.39 & 91.26 & 89.68~/~90.06 & 90.27 \\
        \cmidrule{2-6}
    	& ET-OOD                 &  22.79&  96.27 & 96.26~/~95.53  & 93.65  \\
            & ET-OOD+                 & 24.94 & 95.79 & 95.79~/~94.69 & 92.58 \\
        \cmidrule{2-6}
        & PSA  w/o retraining               & 19.00 & 96.74 & 96.48~/~96.45 & 94.42 \\
        & PSA              &  \textbf{13.06}  & \textbf{97.52} & \textbf{97.47}~/~\textbf{96.95} & \textbf{94.45} \\
    \midrule
    \midrule
     \multirow{6}{*}{\rotatebox{90}{CIFAR-100+TIN}} 	
     & UDG                   &  75.45  & 79.63  & 80.69~/~74.10  & 67.38 \\
        & UDG+                  &  72.84 & 81.64  & 82.06~/~76.94  & \textbf{73.74} \\
        \cmidrule{2-6}
    	& ET-OOD                & 76.90  & 81.96 & 83.54~/~75.63 & 70.84 \\
        & ET-OOD+                 &  78.23 & 81.43 & 84.13~/~73.02 & 72.05\\
        \cmidrule{2-6}
        & PSA  w/o retraining                 &  67.59  & 84.49 & 85.21~/~79.45 & 73.60\\
        & PSA               &  \textbf{60.37} & \textbf{84.53} & \textbf{85.43~/~80.20} & 72.98 \\
     
     \bottomrule
    \end{tabular}
    }
\end{table} 

\textbf{Influence of OOD scoring functions.}
Following \cite{lu2023uncertainty}, we test our method under three different OOD scoring functions (MSP, Energy, T-Energy) and compare it with UDG and ET-OOD, the results are shown in Table ~\ref{T:metrics}. It can be seen that our method achieves the best results in two SCOOD benchmarks under all scoring functions, which strongly proves the universality of our method. By selecting more reliable auxiliary ID/OOD samples by PSA and fully training them, the model can better learn to discriminate ID/OOD samples according to high-level semantic information, so that the model can adapt to various OOD score functions.

\begin{table}[t]
\centering
\renewcommand{\arraystretch}{1.}
\renewcommand{\tabcolsep}{3.pt}
\small
    \caption{\textbf{Influence of different OOD scoring functions.} }
    \label{T:metrics}
    \centering
    \begin{tabular}{ccc|ccc}
    	\toprule
    	 $\mathcal{D}$ & OOD Score & Method & \footnotesize{FPR95~$\downarrow$} & \footnotesize{AUROC~$\uparrow$} & \footnotesize{AUPR-IN/OUT~$\uparrow$} \\
    	\midrule
            \multirow{9}{*}{\rotatebox{90}{CIFAR10+TIN}}
    	& \multirow{3}{*}{MSP}        
    	& UDG\cite{yang2021semantically}                   & 36.22 & 93.78 & 92.61~/~92.94\\
           & & ET-OOD \cite{lu2023uncertainty}                   &  22.79&  96.27 & 96.26~/~95.53\\
            & & PSA                 & \textbf{13.06} & \textbf{97.52} & \textbf{97.47~/~96.95} \\
            \cmidrule{2-6}
    	& \multirow{3}{*}{Energy}        
    	& UDG\cite{yang2021semantically}      & 34.90&  90.65 & 91.56~/~91.11 \\
        & & ET-OOD\cite{lu2023uncertainty}      &  12.86 &   96.05 &  97.04~/~95.01\\
    	& & PSA    & \textbf{10.79} & \textbf{97.74} & \textbf{97.71~/~97.13} \\
            \cmidrule{2-6}
    	& \multirow{3}{*}{T-Energy}        
    	& UDG\cite{yang2021semantically}   &  21.57 &  92.44 & 91.70~/~92.12 \\
        & & ET-OOD\cite{lu2023uncertainty}      & 8.53 & 96.47 & 97.10~/~95.65 \\
    	& & PSA                 &  \textbf{7.60} & \textbf{97.68} & \textbf{97.65~/~97.03} \\

     \midrule
     \midrule
     \multirow{9}{*}{\rotatebox{90}{CIFAR100+TIN}}
    	& \multirow{3}{*}{MSP}        
    	& UDG\cite{yang2021semantically}                   & 75.45  & 79.63 & 80.69~/~74.10 \\
           & & ET-OOD \cite{lu2023uncertainty}                    & 76.90  & 81.96 & 83.54~/~75.63\\
            & & PSA                 & \textbf{60.37} & \textbf{84.53} & \textbf{85.43~/~80.20}\\
            \cmidrule{2-6}
    	& \multirow{3}{*}{Energy}        
    	& UDG\cite{yang2021semantically}       & 76.27 & 79.37 &  79.14~/~74.98\\
        & & ET-OOD\cite{lu2023uncertainty}       & 74.16  & 82.68 & 84.49~/~75.46\\
    	& & PSA     & \textbf{70.69} & \textbf{83.82} & \textbf{86.04~/~76.27} \\
            \cmidrule{2-6}
    	& \multirow{3}{*}{T-Energy}        
    	& UDG\cite{yang2021semantically}    & 74.59 & 79.38 & 79.05~/~75.11 \\
        & & ET-OOD\cite{lu2023uncertainty}       & 41.05 &  82.44  &  84.37~/~76.47\\
    	& & PSA                  & \textbf{32.74} & \textbf{85.14} &\textbf{85.94~/~80.75} \\
    	\bottomrule
    \end{tabular}
\end{table}

\begin{table}[t]
\color{black}
\centering
\renewcommand{\arraystretch}{1.}
\renewcommand{\tabcolsep}{4.pt}
\small
    \caption{ \textbf{The results of different threshold strategies.}}\label{tab:thres_method}
    \resizebox{1.0\columnwidth}{!}{
    \begin{tabular}{c|cccc}
    	\toprule
    	 Method & FPR95~$\downarrow$ & \footnotesize{AUROC~$\uparrow$} & \footnotesize{AUPR-IN/OUT~$\uparrow$} & \footnotesize{ACC~$\uparrow$} \\
    	\midrule
    	Softmax+                   & 14.93$\pm$0.3 & 97.14$\pm$0.03 & 96.92$\pm$0.05~/~96.65$\pm$0.04  & 93.96$\pm$0.02 \\
    	Sort+                 & 18.58$\pm$0.5 & 96.42$\pm$0.04& 95.83$\pm$0.03~/~95.96$\pm$0.07  & 93.62$\pm$0.04\\
        PSA                & \textbf{12.90$\pm$0.2} & \textbf{97.52$\pm$0.04} & \textbf{97.45$\pm$0.06~/~96.96$\pm$0.02} & \textbf{94.42$\pm$0.03}  \\
    	\bottomrule
    \end{tabular}}
\end{table} 

\begin{table}[ht]
\color{black}
\setlength{\tabcolsep}{1pt}
\small
\caption{ \textbf{The results with different $q_{id}$ and  $q_{ood}$ on CIFAR10-benchmark.}}\label{tab:threshold}
\centering
\resizebox{\columnwidth}{!}{
\begin{tabular}{ l c| c | c| c| c| c | c| c |c| c }

\hline
\multicolumn{2}{ l| }{$q_{id}$ \textbackslash{} $q_{ood}$} & 0 & 0.01 & 0.02 & 0.05 & 0.1 & 0.2 & 0.3 & 0.4 & 0.5  \\
\hline
\multirow{2}{*}{0.6}  &FPR95~$\downarrow$ & 66.60 & 67.95 & 87.35 & 63.62 & 85.33 & 22.00 & 21.18 & 16.72  & 17.39  \\
 & AUROC~$\uparrow$ & 83.99  & 81.49 & 73.52  & 81.33 & 74.55 & 95.61 & 95.83 & 96.72 & 96.61   \\
\hline
\multirow{2}{*}{0.8}  &FPR95~$\downarrow$ & 67.87 & 55.28 & 47.81 &36.45 &15.40 & 15.86 & 14.19 & 15.00  & 13.79  \\
 & AUROC~$\uparrow$ & 84.04 & 87.96 & 90.19 & 92.64 &97.07 & 97.02 & 97.28 & 97.09 & 97.37   \\
 \hline
 \multirow{2}{*}{0.9}  &FPR95~$\downarrow$ & 49.07 & 34.88 & 23.93 & 19.34 &13.69  & 13.59 & \textbf{13.06} & 15.48 & 14.19  \\
 & AUROC~$\uparrow$ & 91.98 & 94.43 & 96.09 & 96.66 & 97.46 & 97.47 & \textbf{97.52} & 97.27 & 97.37  \\
\bottomrule
\end{tabular}
}

\end{table}

\subsection{Further Analysis}
\textbf{Different threshold strategy.}
Inspired by \cite{yang2021semantically}, we also introduce two threshold strategies that can replace the energy score in PSA. One is the softmax score, which simply sets $\delta_{id}$ and $\delta_{ood}$ to a fixed softmax probability, while $s_i$ is the output softmax probability of an unlabeled sample $x_i$. The other is SORT softmax, which sorts all unlabeled samples according to softmax probability from largest to smallest, then takes the top ($1-q_{id}$)\% samples as auxiliary ID samples, the last $q_{ood}$\% samples as OOD samples, and the remaining samples as unconfident samples. We call the former Softmax and the latter Sort, and the corresponding retraining versions are Softmax+ and Sort+. \revised {We repeat running these experiments on the CIFAR-10 benchmark three times and report the average results with standard deviations in Table~\ref{tab:thres_method}.} It can be seen that PSA based on energy score achieves the best results. Although the latter two also use PSA as a sample assignment strategy for SCOOD training, compared with Softmax+, PSA employs energy score as the threshold strategy, which considers the overall uncertainty of samples, so that the model can choose more reliable auxiliary ID/OOD samples during training. In the meanwhile, Softmax+ still significantly outperforms ET-OOD, which demonstrates the importance of reducing noisy samples by conducting ternary sample assignment during model training. Compared with PSA, Sort+ selects a fixed number of auxiliary ID/OOD samples during training, which makes it easy to select too many or too few samples to participate in training, resulting in inferior OOD performance.

\textbf{Analysis of $q_{i d}$ and $q_{o o d}$.} Here we analyze the impact of threshold values $q_{i d}$ and $q_{o o d}$. Our method is concise and effective, when performing the auxiliary ID and OOD sample selection, only the thresholds $q_{i d}$ and $q_{o o d}$ need to be set. \revised{In general, $q_{id}$ and $q_{ood}$ are two hyper-parameters that affect each other. For simplicity, we just fix one and pick the other. } Table ~\ref{tab:threshold} shows the results of different combinations of  $q_{i d}$ and $q_{o o d}$ on the CIFAR10-benchamrk. It can be seen that stable and good results are achieved when $q_{i d}$ is set to 0.9 and the value of $q_{o o d}$ is in the range of [0.1,0.3]. 

\begin{table}[h]
\color{black}
\setlength{\tabcolsep}{1pt}
\centering
\small
\caption{\textbf{The results with different $T_{warm}$ on CIFAR-10 benchmark.}}\label{tab:warm_epoch}
\begin{tabular}{c|c|c|c|c|c|c|c}
\hline
$T_{warm}$ & 5 & 10 & 20 & 30 & 50 & 80 & 100 \\
\hline
FPR95~$\downarrow$ & 14.85 & 15.12 & 14.09 & 13.06 & 14.01 & 15.86 & 17.52 \\
AUROC~$\uparrow$ & 97.38 & 97.33 & 97.35 & 97.52  & 97.39 & 97.27  & 97.19 \\
\bottomrule
\end{tabular}
\end{table}

\revised{\textbf{Analysis of $T_{warm}$.}
We conducted an analysis of the impact of $T_{warm}$, Table \ref{tab:warm_epoch} reports the results of different values of $T_{warm}$ on the CIFAR-10 benchmark, and it can be seen that stable and good results can be achieved when $T_{warm}$ is in the range of [20,50].}

\textbf{Analysis of $\lambda$.} $\lambda$ is the the balancing weight of  concept contrastive learning loss $\mathcal{L}_{\mathrm{CCL}}$. As shown in Table \ref{tab:loss_weight}, Our method is robust when $\lambda$ values are in the range [0.1,1], we set it to 0.1 for all experiments.



\begin{table}[h]
\setlength{\tabcolsep}{1pt}
\centering
\small
\caption{\textbf{The results with different $\lambda$ on CIFAR-10 benchmark.}}\label{tab:loss_weight}
\begin{tabular}{c|c|c|c|c|c|c}
\hline
$\lambda$ & 0.1 & 0.2 & 0.5 & 1 & 1.5 & 2 \\
\hline
FPR95~$\downarrow$ & 13.06 & 13.00 & 12.88 & 13.35 & 13.75 & 13.96 \\
AUROC~$\uparrow$ & 97.52 & 97.52 & 97.52  & 97.45 & 97.46  & 97.38 \\
\bottomrule
\end{tabular}
\end{table}

\textbf{Effectiveness of CCL.} We analyze the performance of different auxiliary contrastive representation learning losses $\mathcal{L}_{\text {Aux }}$ applied to PSA, including self-supervised contrastive learning (SSL), supervised contrastive learning \cite{khosla2020supervised} (SCL) and our concept contrastive representation learning (CCL). SSL performs infoNCE \cite{oord2018representation} loss on all training data so that the representations of different views of the same image are close together in the embedding space. SCL performs supervised contrastive learning on the labeled dataset so that images of the same semantic class are close together in the embedding space, while images of different classes are far apart as possible in the embedding space. However, our CCL explicitly separates ID samples from selected OOD samples with different semantic concepts in the embedding space. \revised {We repeat running these experiments on the CIFAR-10 benchmark three times and report the average results with standard deviations in Table~\ref{T:Rep}. As can be seen from Table~\ref{T:Rep},} PSA achieves the best results when using CCL, which confirms the superiority of explicitly widening the distance between ID and OOD samples in the representation space.  

\begin{table}[t]
\color{black}
\centering
\renewcommand{\arraystretch}{1.}
\renewcommand{\tabcolsep}{4.pt}
\small
    \caption{\textbf{The design choice of $L_{Aux }$.} SSL stands for self-supervised contrastive learning loss. SCL stands for supervised contrastive learning loss. CCL stands for the proposed concept contrastive learning loss.  }
    \label{T:Rep}
    \resizebox{\columnwidth}{!}{
    \begin{tabular}{c|cccc}
    	\toprule
    	 $L_{Aux }$ & FPR95~$\downarrow$ & \footnotesize{AUROC~$\uparrow$} & \footnotesize{AUPR-IN/OUT~$\uparrow$} & \footnotesize{ACC~$\uparrow$} \\
    	\midrule
    	SSL                   & 14.48$\pm$0.2 & 97.29$\pm$0.04 & 97.26$\pm$0.07~/~96.73$\pm$0.04  & 93.94$\pm$0.08 \\
    	SCL                 & 15.75$\pm$0.1 & 97.23$\pm$0.01 & 97.18$\pm$0.01~/~96.58$\pm$0.03  & 93.97$\pm$0.02\\
        CCL                 & \textbf{12.90$\pm$0.2} & \textbf{97.52$\pm$0.04} & \textbf{97.45$\pm$0.06~/~96.96$\pm$0.02} & \textbf{94.42$\pm$0.03}  \\
    	\bottomrule
    \end{tabular}}
\end{table}

\begin{table}[t]
\color{black}
      \small
      \centering
      \caption{The comparison with diffusion model-based method.}
      \begin{tabular}{lcccc}
        \toprule
        \multirow{2}[3]{*}{Method}  &
        \multicolumn{4}{c}{CIFAR-10 benchmark} \\
        \cmidrule(lr){2-5} 
         & SVHN  & CelebA & VFlip & HFlip \\
        \midrule
        DDMP\cite{Graham_2023_CVPR} & 97.9 & 68.5 & 63.2 & 50.5 \\
        PSA & \textbf{99.9} & \textbf{99.1} & \textbf{96.7}  & \textbf{78.8}  \\
        \bottomrule
      \end{tabular}
      \label{tab:DDPM}
\end{table}

\revised{\textbf{Comparison with diffusion model-based method.} 
We also compare with DDPM\cite{Graham_2023_CVPR}, which is a recently proposed reconstruction-based OOD detection method. DDPM reconstructs the image disturbed by noise by using denoising diffusion probabilistic models \cite{ho2020denoising}, and uses the value of reconstruction error for OOD detection. Since its setup is quite different from SCOOD, We tested our method under the CIFAR-10 benchmark they used, and Table \ref{tab:DDPM} reports the comparison results under the AUC score metric. As can be seen from Table \ref{tab:DDPM}, our method can achieve better results compared to DDPM.}

\begin{table}[t]
\color{black}
\centering
\renewcommand{\arraystretch}{1.}
\renewcommand{\tabcolsep}{4.pt}
\small
    \caption{ \textbf{The results of different training strategies.}}\label{tab:joint}
    \begin{tabular}{c|cccc}
    	\toprule
    	 Method & FPR95~$\downarrow$ & \footnotesize{AUROC~$\uparrow$} & \footnotesize{AUPR-IN/OUT~$\uparrow$} & \footnotesize{ACC~$\uparrow$} \\
    	\midrule
    	PSA (Joint)               & 13.23 & 97.46 & 97.34/96.99  & 94.45 \\
        PSA                & \textbf{13.06} & \textbf{97.52} & \textbf{97.47/96.95} & \textbf{94.45}  \\
    	\bottomrule
    \end{tabular}
\end{table} 

\revised{\textbf{Joint training strategy for PSA.} To integrating retraining to the main training process, we can use CosineAnnealingWarmRestarts as the learning rate annealing strategy, which can make the learning rate return to the initial value periodically. Therefore, we can set the iteration period of the learning rate to 2, and perform sample selection training in the first learning rate decay period and retraining in the second learning rate decay period. Table \ref{tab:joint} reports the results of this joint training strategy. It can be seen from the table that our proposed joint training strategy can achieve results that are very close to the naive PSA. }

\begin{table}[t]
\color{black}
\centering
\renewcommand{\arraystretch}{1.}
\renewcommand{\tabcolsep}{4.pt}
\small
    \caption{ \textbf{The results of different concept contrastive loss.}}\label{tab:MCCL}
    \begin{tabular}{c|cccc}
    	\toprule
    	 Method & FPR95~$\downarrow$ & \footnotesize{AUROC~$\uparrow$} & \footnotesize{AUPR-IN/OUT~$\uparrow$} & \footnotesize{ACC~$\uparrow$} \\
    	\midrule
        MCCL (n=5)	               & 13.65 & 97.43& 97.29/96.87  & 94.42 \\
         MCCL (n=10)	               & 13.55 & 97.40& 97.26/96.87  & 94.35 \\
          MCCL (n=20)	               & 13.82 & 97.43& 97.34/96.92  & 94.42 \\
        CCL                & 
        \textbf{13.06} & \textbf{97.52} & \textbf{97.47/96.95} & \textbf{94.45}  \\
    	\bottomrule
    \end{tabular}
\end{table}

\revised{\textbf{Whether selected OOD samples need to be grouped according to their semantics to help CCL explore subtle differences between them?} To cluster the selected OOD samples into multiple semantic groups, at the end of each epoch, we use $k$-means to cluster the selected OOD samples into $n$ clusters and treat each cluster as a separate semantic concept to construct the multiple concept contrastive loss (MCCL). Table \ref{tab:MCCL} reports the results of MCCL when the values of $n$ are 5, 10, 20, respectively. It can be seen that there is a slight performance decrease compared with CCL, this is because the goal of the OOD detection task is to separate ID and OOD samples. Compared with dividing OOD samples into multiple semantic concepts, treating all OOD samples as a simple OOD concept can better separate ID samples and OOD samples in the representation space. }

\begin{figure}[t]
  \setlength{\tabcolsep}{0.5mm}
  \small
  \centering
  \begin{tabular}{cc}
    (a) \small ID Sample Filtered  & (b) Clean ID Sample Filtered \\
    \includegraphics[width=0.23\textwidth,height=0.18\textwidth]{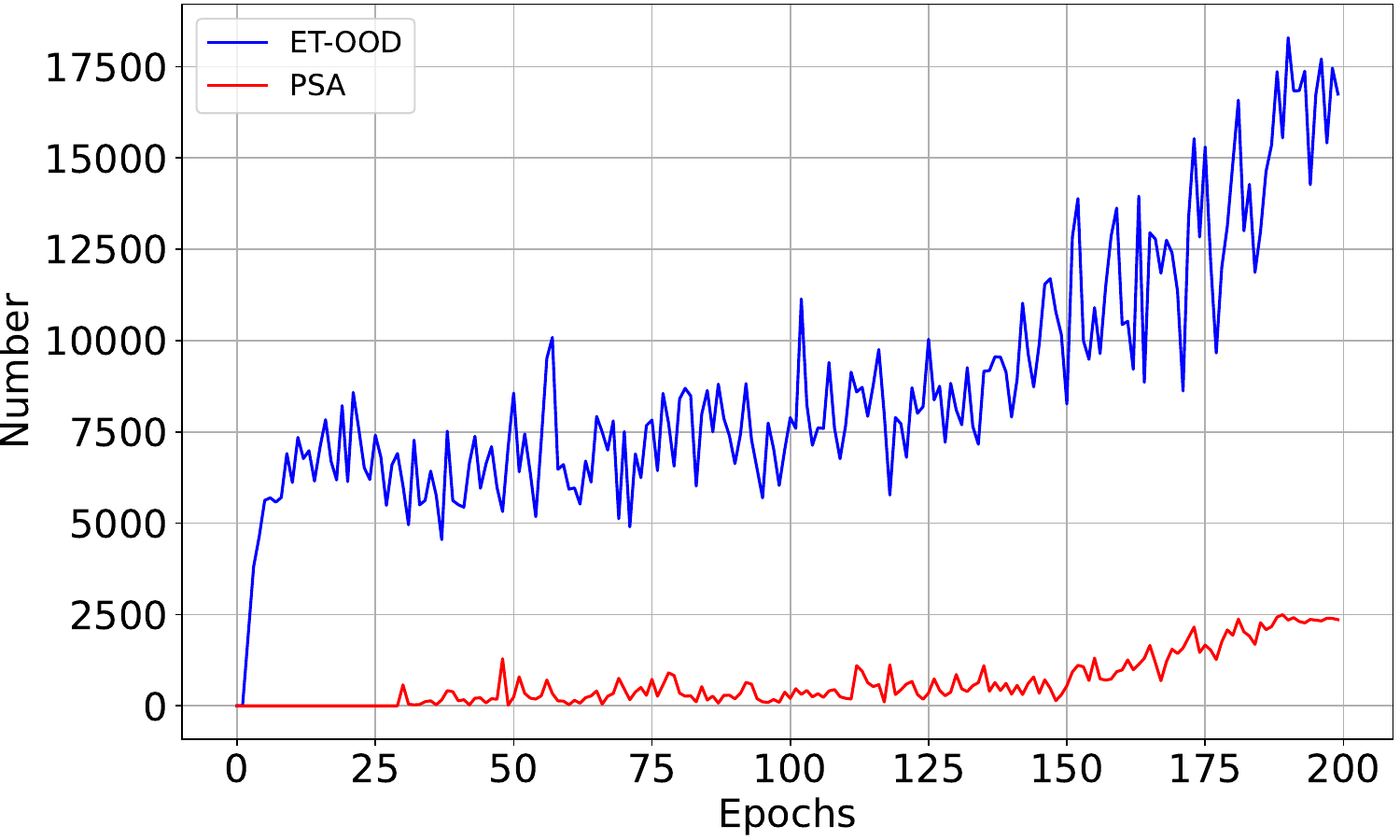}
    
    & 
    \includegraphics[width=0.23\textwidth,height=0.18\textwidth]{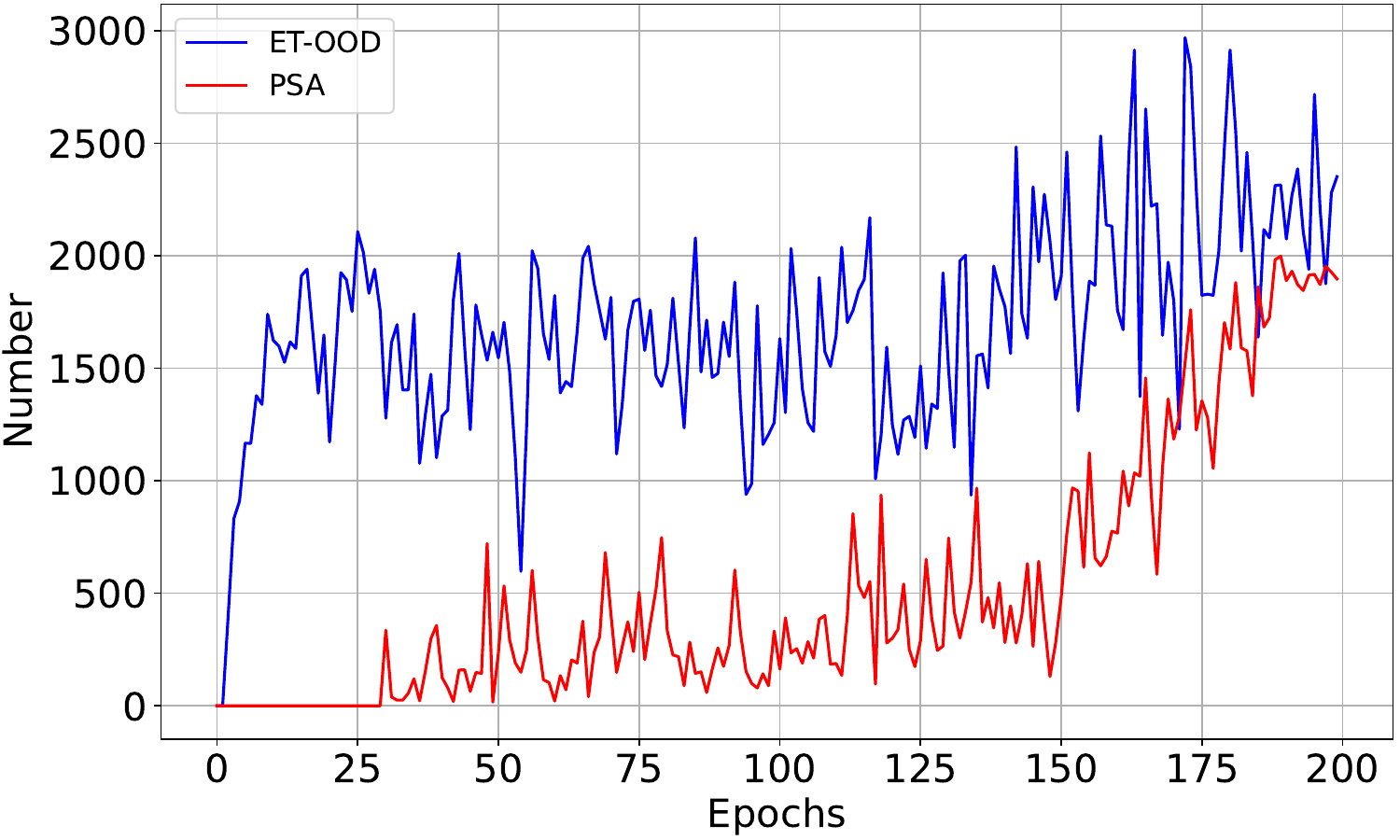}
    
    \\
    (c) \small OOD Sample Filtered & (d) Clean OOD Sample Filtered    \\
    \includegraphics[width=0.23\textwidth,height=0.18\textwidth]{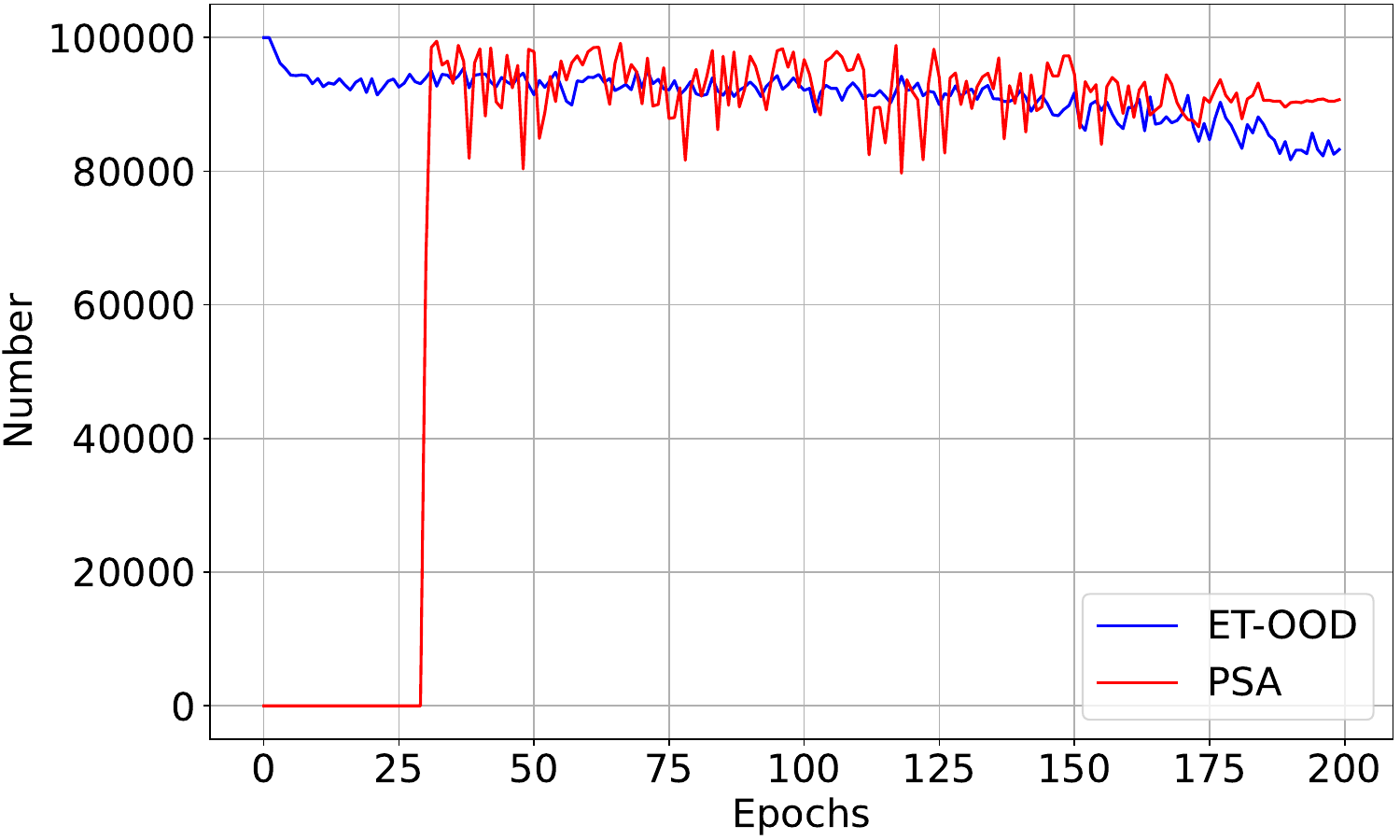}

    & 
    \includegraphics[width=0.23\textwidth,height=0.18\textwidth]{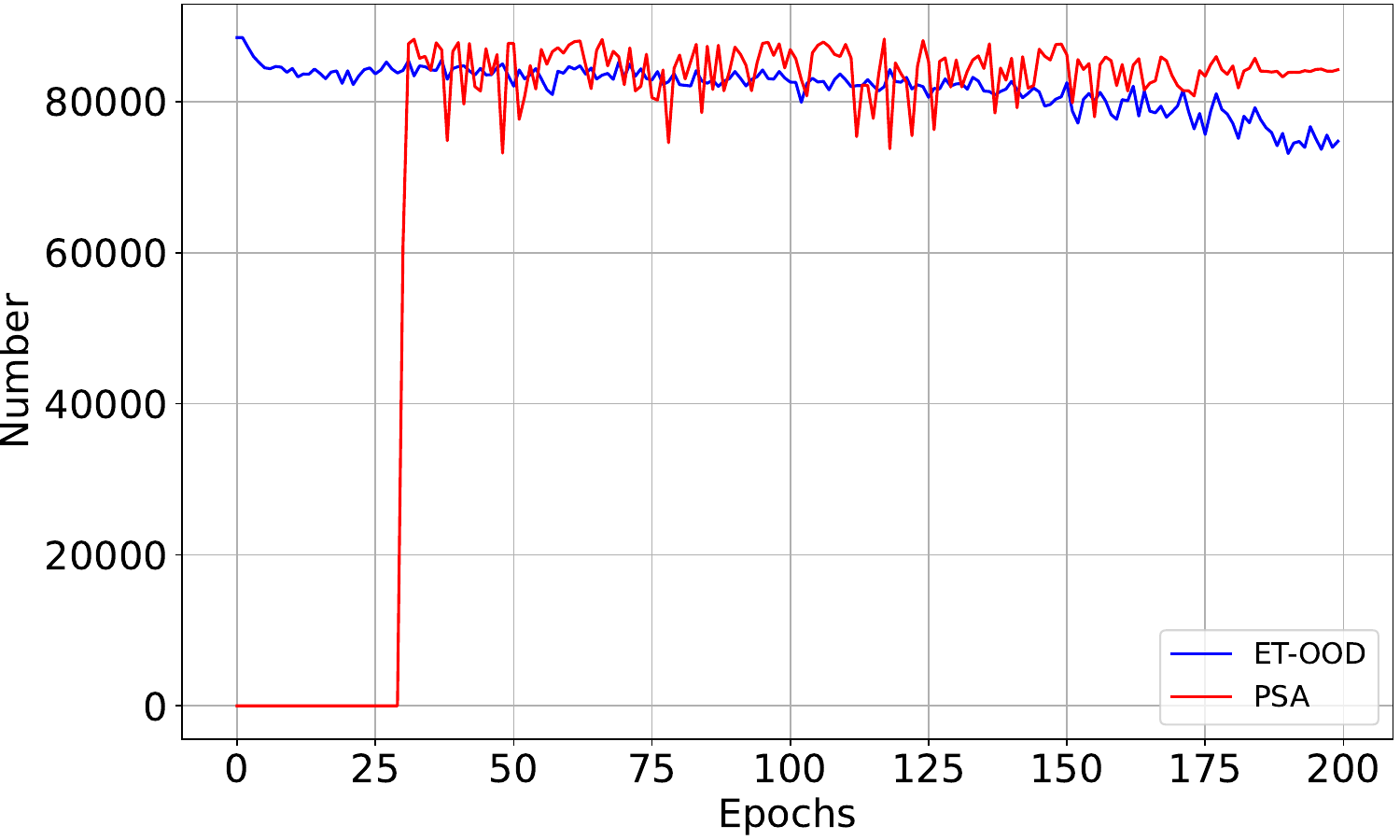}
    
    \\
    \end{tabular}
 \caption{The number of selected samples during training. (a) and (b) show that the number of clean ID samples selected by PSA is slightly lower than ET-OOD, but  the purity of selected ID samples is significantly higher than ET-OOD. (c) and (d) show that both the quantity and purity of clean OOD samples selected by PSA are higher than that of ET-OOD.}
\label{fig:sample_nums}
\end{figure}

\begin{figure}
\color{black}
    \centering    \includegraphics[width=0.8\columnwidth]{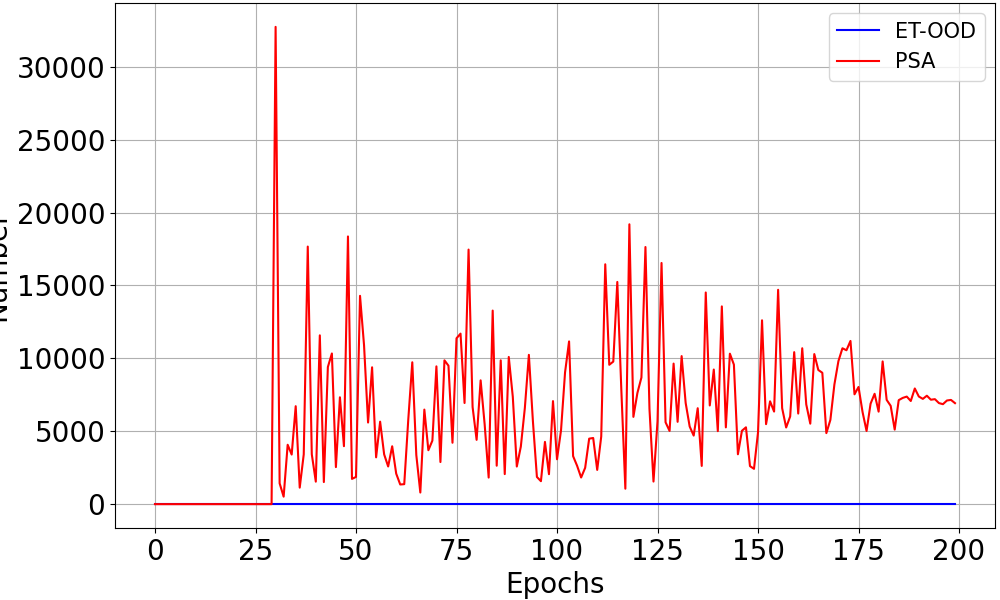}
    \vspace{-0.3cm}
    \caption{The number of selected unconfident samples during training.}
    \label{fig:discard}
\end{figure}

\begin{figure}
    \centering
    \includegraphics[width=0.99\columnwidth]{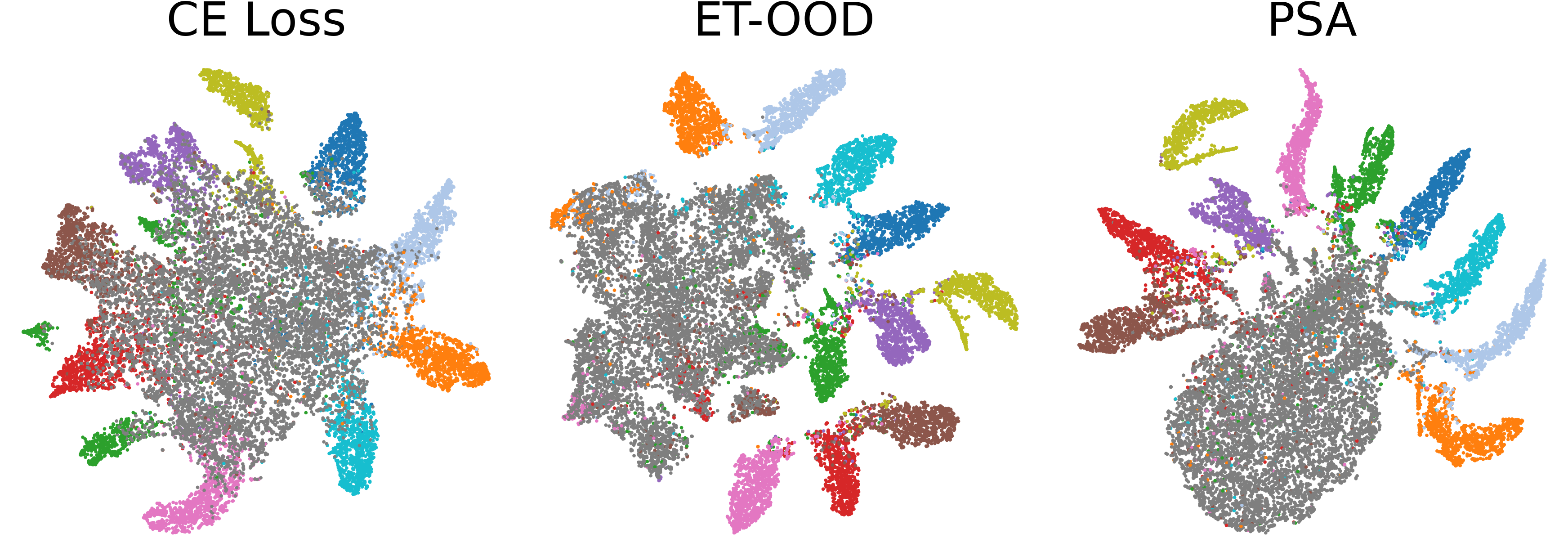}
    \vspace{-0.3cm}
    \caption{t-SNE visualisation of the representations of ID
classes (colored) data and all OOD (dark grey) data on CIFAR-10 benchmark.}
    \label{fig:t-sne}
\end{figure}

\textbf{The analysis of selected samples.} We visualize the change in the number of auxiliary ID and OOD samples selected by PSA when training on the CIFAR-10 benchmark, the results compared with ET-OOD are presented in Fig. \ref{fig:sample_nums}. From (a) and (b), it can be seen that there is a small difference between the number of clean ID samples selected by PSA and ET-OOD at the end of training (1897 vs. 2351), but the purity of auxiliary ID samples selected by our PSA is much higher than ET-OOD (80\% vs. 14\%). From (c) and (d), it can be seen that the PSA not only selects higher purity of auxiliary OOD samples than ET-OOD (92.9\% vs. 89.8\%), but also PSA selects more clean OOD samples (84235 vs. 74757), which confirms that our method can indeed select more reliable auxiliary ID/OOD samples from unlabeled dataset. In addition, from (a) and (b), we can see that PSA selects a large number of auxiliary ID samples only in the late stage of model training, which makes it difficult for the model to fit them well, confirming the importance of retraining strategy. \revised{In addition, in Fig. \ref{fig:discard}, we also visualize the number of unconfident samples discarded when PSA is trained on the CIFAR-10 benchmark as the training epoch increases. As can be seen from the figure, as the training proceeds, the number of unconfident samples is non-trivial and gradually stable. In contrast, the number of unconfident samples in ET-OOD remains 0 all the time, making the model inevitably fit noisy samples, leading to suboptimal performance.}  

\begin{table}[t]
\color{black}
      \small
      \centering
      \caption{The comparison of running time.}
      \begin{tabular}{lcccc}
        \toprule
        \multirow{2}[3]{*}{Method}  &
        \multicolumn{2}{c}{CIFAR-10 benchmark} & \multicolumn{2}{c}{CIFAR-100 benchmark}\\
        \cmidrule(lr){2-3} \cmidrule(lr){4-5}
         & Training & Inference & Training & Inference \\
        \midrule
        ET-OOD & 12.7h & 8.9m & 12.9h & 8.9m \\
        PSA & 8.1h & 8.9m & 8.2h  & 8.9m  \\
        \bottomrule
      \end{tabular}
      \label{tab:cost_time}
\end{table}

\revised{\textbf{Computational complexity analysis.} 
Our PSA method is trained with a convolutional neural network, which mainly contains convolutional layers, activation functions and a pooling layer. The training mode is batch processing, and the loss functions are cross-entropy loss and supervised contrastive learning loss. The computational complexity of these components scales linearly with the size of input data. Thus the computational complexity of our method is rough linear with the size of training data.
For the running time, although PSA requires an additional retraining process, the energy score is computed by a linear classifier. Compared with clustering-based ID sample filtering strategy, the computational complexity of the energy-based ternary sample assignment strategy is significant lower. We report the running time of our method and ET-OOD on the two SCOOD benchmarks in Table \ref{tab:cost_time}. As can be seen from Table \ref{tab:cost_time}, the overall running time of our method is less compared to ET-OOD.} 

\begin{figure}
\color{black}
    \centering    \includegraphics[width=0.99\columnwidth]{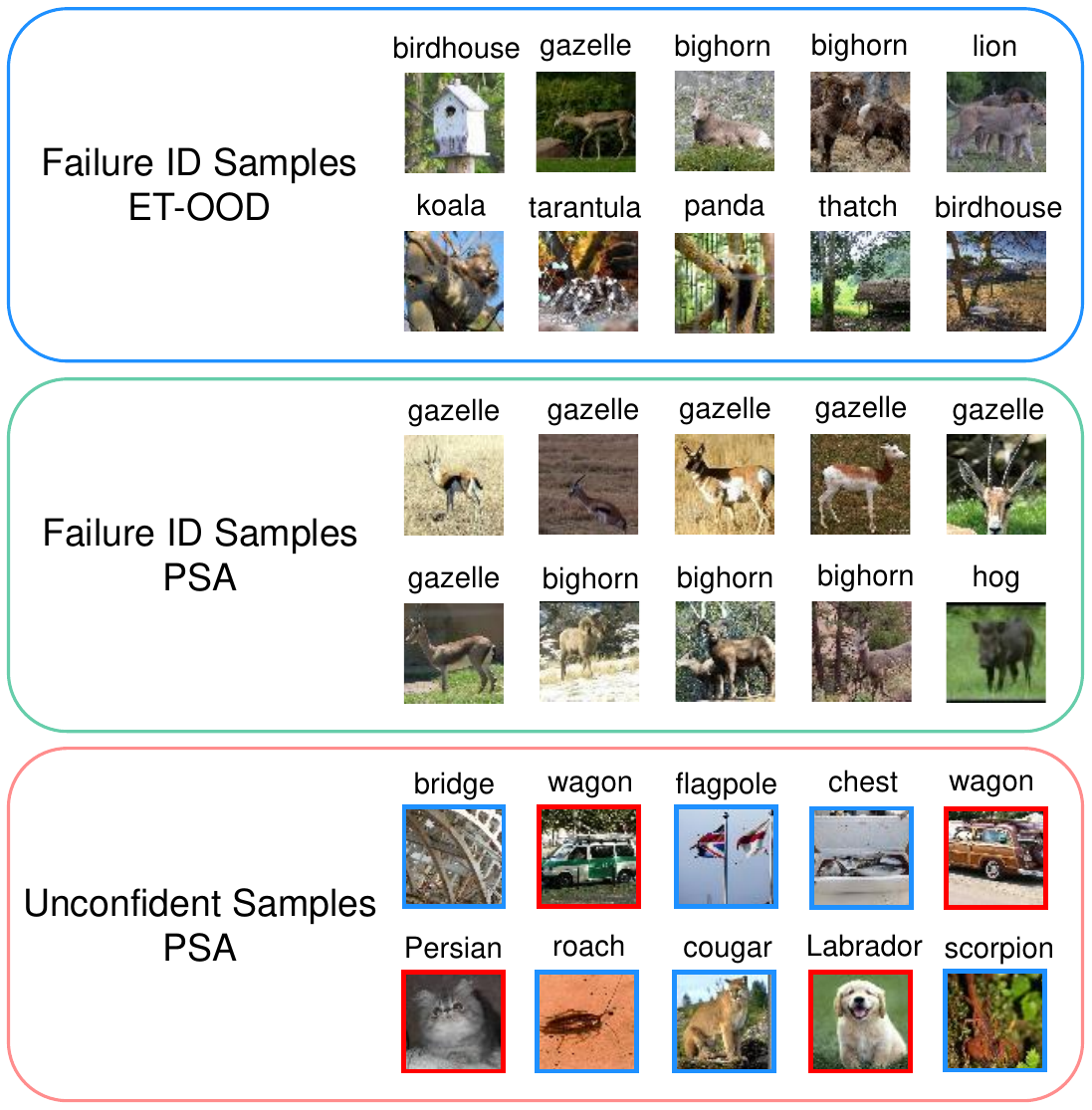}
    \caption{The visualisation of images selected by ET-OOD and PSA. }
    \label{fig:selected}
\end{figure}

\textbf{Visualization results.} 
In addition to the quantitative results, we also perform a qualitative analysis on CIFAR-10 benchmark. In Fig. \ref{fig:t-sne}, we visualize the feature space learned by the model with t-SNE \cite{van2008visualizing}. From the figure, we can observe that training the model only on the labeled ID dataset with cross-entropy loss will cause the ID and OOD data to be entangled in the representation space and difficult to distinguish. Although ET-OOD based on IDF strategy makes the identification boundary of ID and OOD samples less blurred, there is still a large number of entanglement between ID and OOD samples. In contrast, our PSA generates a more compact representation of the ID and OOD samples and keeps them well separated in the representation space. \revised{In addition to the feature space, we also visualize the selected images in Fig. \ref{fig:selected}. For the failure ID samples, we randomly selected 10 images from the class ‘‘deer". As can be seen from the figure, compared with ET-OOD, the failure ID samples selected by our method are semantically closer to the ‘‘deer" class, confirming the better semantically discrimination of our method. For the selected unconfident samples, we also randomly selected 10 images for visualization. As can be seen from the figure, a high proportion of ID class (red box) and OOD class (blue box) auxiliary samples simultaneously exist in the selected unconfident samples. Our method excludes them from the model training, thus reducing the fitting to noise and improving the model performance.}

\revised{\textbf{The disadvantages of the proposed method.}
Our method requires an additional retraining step, which may lead to some tedious on
practical deployment. However, this trouble could be effectively solved by the proposed joint
training strategy.}

\section{Conclusions}
In this paper, we propose a predictive sample assignment (PSA) framework to effectively detect OOD samples in SCOOD benchmarks by learning semantic discriminative representation. We point out that the current ID sample filtering strategy based on binary decision leads to a lot of noise, resulting in inferior performance. By introducing a dual-threshold based ternary sample assignment strategy, PSA can identify uncertain samples and exclude them from training, leading to the selection of higher purity auxiliary ID/OOD samples. Extensive experiments on two SCOOD benchmarks show that PSA achieves state-of-the-art results. 

\bibliographystyle{IEEEtran}

\bibliography{reference}

@string(CVPR  = "{IEEE Conf. Comput. Vis. Pattern Recog.}")

@string(ECCV  = "{Eur. Conf. Comput. Vis.}")

@string(ECCV = "{ECCV}")

@string(CVPR = "{CVPR}")

@inproceedings{yang2021semantically,
  title={Semantically coherent out-of-distribution detection},
  author={Yang, Jingkang and Wang, Haoqi and Feng, Litong and Yan, Xiaopeng and Zheng, Huabin and Zhang, Wayne and Liu, Ziwei},
  booktitle={Proceedings of the IEEE/CVF International Conference on Computer Vision},
  pages={8301--8309},
  year={2021}
}

@inproceedings{lu2023uncertainty,
  title={Uncertainty-Aware Optimal Transport for Semantically Coherent Out-of-Distribution Detection},
  author={Lu, Fan and Zhu, Kai and Zhai, Wei and Zheng, Kecheng and Cao, Yang},
  booktitle={Proceedings of the IEEE/CVF Conference on Computer Vision and Pattern Recognition},
  pages={3282--3291},
  year={2023}
}

@article{liu2020energy,
  title={Energy-based out-of-distribution detection},
  author={Liu, Weitang and Wang, Xiaoyun and Owens, John and Li, Yixuan},
  journal={Advances in neural information processing systems},
  volume={33},
  pages={21464--21475},
  year={2020}
}

@inproceedings{cimpoi2014describing,
  title={Describing textures in the wild},
  author={Cimpoi, Mircea and Maji, Subhransu and Kokkinos, Iasonas and Mohamed, Sammy and Vedaldi, Andrea},
  booktitle={Proceedings of the IEEE conference on computer vision and pattern recognition},
  pages={3606--3613},
  year={2014}
}

@article{netzer2011reading,
  title={Reading digits in natural images with unsupervised feature learning},
  author={Netzer, Yuval and Wang, Tao and Coates, Adam and Bissacco, Alessandro and Wu, Bo and Ng, Andrew Y},
  year={2011}
}

@article{le2015tiny,
  title={Tiny imagenet visual recognition challenge},
  author={Le, Ya and Yang, Xuan},
  journal={CS 231N},
  volume={7},
  number={7},
  pages={3},
  year={2015}
}

@article{yu2015lsun,
  title={Lsun: Construction of a large-scale image dataset using deep learning with humans in the loop},
  author={Yu, Fisher and Seff, Ari and Zhang, Yinda and Song, Shuran and Funkhouser, Thomas and Xiao, Jianxiong},
  journal={arXiv preprint arXiv:1506.03365},
  year={2015}
}

@article{zhou2017places,
  title={Places: A 10 million image database for scene recognition},
  author={Zhou, Bolei and Lapedriza, Agata and Khosla, Aditya and Oliva, Aude and Torralba, Antonio},
  journal={IEEE transactions on pattern analysis and machine intelligence},
  volume={40},
  number={6},
  pages={1452--1464},
  year={2017},
  publisher={IEEE}
}

@article{liang2017enhancing,
  title={Enhancing the reliability of out-of-distribution image detection in neural networks},
  author={Liang, Shiyu and Li, Yixuan and Srikant, Rayadurgam},
  journal={arXiv preprint arXiv:1706.02690},
  year={2017}
}

@article{hendrycks2018deep,
  title={Deep anomaly detection with outlier exposure},
  author={Hendrycks, Dan and Mazeika, Mantas and Dietterich, Thomas},
  journal={arXiv preprint arXiv:1812.04606},
  year={2018}
}

@inproceedings{yu2019unsupervised,
  title={Unsupervised out-of-distribution detection by maximum classifier discrepancy},
  author={Yu, Qing and Aizawa, Kiyoharu},
  booktitle={Proceedings of the IEEE/CVF international conference on computer vision},
  pages={9518--9526},
  year={2019}
}

@article{hendrycks2016baseline,
  title={A baseline for detecting misclassified and out-of-distribution examples in neural networks},
  author={Hendrycks, Dan and Gimpel, Kevin},
  journal={arXiv preprint arXiv:1610.02136},
  year={2016}
}

@article{lee2018simple,
  title={A simple unified framework for detecting out-of-distribution samples and adversarial attacks},
  author={Lee, Kimin and Lee, Kibok and Lee, Honglak and Shin, Jinwoo},
  journal={Advances in neural information processing systems},
  volume={31},
  year={2018}
}

@techreport{bulusu2020anomalous,
  title={Anomalous instance detection in deep learning: A survey},
  author={Bulusu, Saikiran and Kailkhura, Bhavya and Li, Bo and Varshney, P and Song, Dawn},
  year={2020},
  institution={Lawrence Livermore National Lab.(LLNL), Livermore, CA (United States)}
}

@inproceedings{nguyen2015deep,
  title={Deep neural networks are easily fooled: High confidence predictions for unrecognizable images},
  author={Nguyen, Anh and Yosinski, Jason and Clune, Jeff},
  booktitle={Proceedings of the IEEE conference on computer vision and pattern recognition},
  pages={427--436},
  year={2015}
}

@article{choi2018generative,
  title={Generative ensembles for robust anomaly detection},
  author={Choi, Hyunsun and Jang, Eric},
  year={2018}
}

@inproceedings{geiger2012we,
  title={Are we ready for autonomous driving? the kitti vision benchmark suite},
  author={Geiger, Andreas and Lenz, Philip and Urtasun, Raquel},
  booktitle={2012 IEEE conference on computer vision and pattern recognition},
  pages={3354--3361},
  year={2012},
  organization={IEEE}
}

@article{huang2021importance,
  title={On the importance of gradients for detecting distributional shifts in the wild},
  author={Huang, Rui and Geng, Andrew and Li, Yixuan},
  journal={Advances in Neural Information Processing Systems},
  volume={34},
  pages={677--689},
  year={2021}
}

@inproceedings{sun2022out,
  title={Out-of-distribution detection with deep nearest neighbors},
  author={Sun, Yiyou and Ming, Yifei and Zhu, Xiaojin and Li, Yixuan},
  booktitle={International Conference on Machine Learning},
  pages={20827--20840},
  year={2022},
  organization={PMLR}
}

@article{sehwag2021ssd,
  title={Ssd: A unified framework for self-supervised outlier detection},
  author={Sehwag, Vikash and Chiang, Mung and Mittal, Prateek},
  journal={arXiv preprint arXiv:2103.12051},
  year={2021}
}

@article{ming2022exploit,
  title={How to exploit hyperspherical embeddings for out-of-distribution detection?},
  author={Ming, Yifei and Sun, Yiyou and Dia, Ousmane and Li, Yixuan},
  journal={arXiv preprint arXiv:2203.04450},
  year={2022}
}

@article{lee2017training,
  title={Training confidence-calibrated classifiers for detecting out-of-distribution samples},
  author={Lee, Kimin and Lee, Honglak and Lee, Kibok and Shin, Jinwoo},
  journal={arXiv preprint arXiv:1711.09325},
  year={2017}
}

@article{vernekar2019out,
  title={Out-of-distribution detection in classifiers via generation},
  author={Vernekar, Sachin and Gaurav, Ashish and Abdelzad, Vahdat and Denouden, Taylor and Salay, Rick and Czarnecki, Krzysztof},
  journal={arXiv preprint arXiv:1910.04241},
  year={2019}
}

@article{du2022vos,
  title={Vos: Learning what you don't know by virtual outlier synthesis},
  author={Du, Xuefeng and Wang, Zhaoning and Cai, Mu and Li, Yixuan},
  journal={arXiv preprint arXiv:2202.01197},
  year={2022}
}

@article{tao2023non,
  title={Non-parametric outlier synthesis},
  author={Tao, Leitian and Du, Xuefeng and Zhu, Xiaojin and Li, Yixuan},
  journal={arXiv preprint arXiv:2303.02966},
  year={2023}
}

@article{wang2024learning,
  title={Learning to augment distributions for out-of-distribution detection},
  author={Wang, Qizhou and Fang, Zhen and Zhang, Yonggang and Liu, Feng and Li, Yixuan and Han, Bo},
  journal={Advances in Neural Information Processing Systems},
  volume={36},
  year={2024}
}

@article{wang2023out,
  title={Out-of-distribution detection with implicit outlier transformation},
  author={Wang, Qizhou and Ye, Junjie and Liu, Feng and Dai, Quanyu and Kalander, Marcus and Liu, Tongliang and Hao, Jianye and Han, Bo},
  journal={arXiv preprint arXiv:2303.05033},
  year={2023}
}

@inproceedings{chen2021atom,
  title={Atom: Robustifying out-of-distribution detection using outlier mining},
  author={Chen, Jiefeng and Li, Yixuan and Wu, Xi and Liang, Yingyu and Jha, Somesh},
  booktitle={Machine Learning and Knowledge Discovery in Databases. Research Track: European Conference, ECML PKDD 2021, Bilbao, Spain, September 13--17, 2021, Proceedings, Part III 21},
  pages={430--445},
  year={2021},
  organization={Springer}
}

@inproceedings{ming2022poem,
  title={Poem: Out-of-distribution detection with posterior sampling},
  author={Ming, Yifei and Fan, Ying and Li, Yixuan},
  booktitle={International Conference on Machine Learning},
  pages={15650--15665},
  year={2022},
  organization={PMLR}
}

@article{lecun2006tutorial,
  title={A tutorial on energy-based learning},
  author={LeCun, Yann and Chopra, Sumit and Hadsell, Raia and Ranzato, M and Huang, Fujie},
  journal={Predicting structured data},
  volume={1},
  number={0},
  year={2006}
}

@article{grathwohl2019your,
  title={Your classifier is secretly an energy based model and you should treat it like one},
  author={Grathwohl, Will and Wang, Kuan-Chieh and Jacobsen, J{\"o}rn-Henrik and Duvenaud, David and Norouzi, Mohammad and Swersky, Kevin},
  journal={arXiv preprint arXiv:1912.03263},
  year={2019}
}

@inproceedings{lin2021mood,
  title={Mood: Multi-level out-of-distribution detection},
  author={Lin, Ziqian and Roy, Sreya Dutta and Li, Yixuan},
  booktitle={Proceedings of the IEEE/CVF conference on Computer Vision and Pattern Recognition},
  pages={15313--15323},
  year={2021}
}

@article{wang2021can,
  title={Can multi-label classification networks know what they don’t know?},
  author={Wang, Haoran and Liu, Weitang and Bocchieri, Alex and Li, Yixuan},
  journal={Advances in Neural Information Processing Systems},
  volume={34},
  pages={29074--29087},
  year={2021}
}

@inproceedings{caron2018deep,
  title={Deep clustering for unsupervised learning of visual features},
  author={Caron, Mathilde and Bojanowski, Piotr and Joulin, Armand and Douze, Matthijs},
  booktitle={Proceedings of the European conference on computer vision (ECCV)},
  pages={132--149},
  year={2018}
}

@article{oord2018representation,
  title={Representation learning with contrastive predictive coding},
  author={Oord, Aaron van den and Li, Yazhe and Vinyals, Oriol},
  journal={arXiv preprint arXiv:1807.03748},
  year={2018}
}

@article{khosla2020supervised,
  title={Supervised contrastive learning},
  author={Khosla, Prannay and Teterwak, Piotr and Wang, Chen and Sarna, Aaron and Tian, Yonglong and Isola, Phillip and Maschinot, Aaron and Liu, Ce and Krishnan, Dilip},
  journal={Advances in neural information processing systems},
  volume={33},
  pages={18661--18673},
  year={2020}
}

@article{van2008visualizing,
  title={Visualizing data using t-SNE.},
  author={Van der Maaten, Laurens and Hinton, Geoffrey},
  journal={Journal of machine learning research},
  volume={9},
  number={11},
  year={2008}
}

@inproceedings{katz2022training,
  title={Training ood detectors in their natural habitats},
  author={Katz-Samuels, Julian and Nakhleh, Julia B and Nowak, Robert and Li, Yixuan},
  booktitle={International Conference on Machine Learning},
  pages={10848--10865},
  year={2022},
  organization={PMLR}
}

@article{fang2024learnability,
  title={On the Learnability of Out-of-distribution Detection},
  author={Fang, Zhen and Li, Yixuan and Liu, Feng and Han, Bo and Lu, Jie},
  journal={Journal of Machine Learning Research},
  volume={25},
  year={2024},
  publisher={MICROTOME PUBL 31 GIBBS ST, BROOKLINE, MA 02446 USA}
}

@article{zhu2024diversified,
  title={Diversified outlier exposure for out-of-distribution detection via informative extrapolation},
  author={Zhu, Jianing and Geng, Yu and Yao, Jiangchao and Liu, Tongliang and Niu, Gang and Sugiyama, Masashi and Han, Bo},
  journal={Advances in Neural Information Processing Systems},
  volume={36},
  year={2024}
}

@article{FangQXL24,
  author       = {Jianwu Fang and
                  Jiahuan Qiao and
                  Jianru Xue and
                  Zhengguo Li},
  title        = {Vision-Based Traffic Accident Detection and Anticipation: {A} Survey},
  journal      = {IEEE Transactions on Circuits and Systems for Video Technology},
  volume       = {34},
  number       = {4},
  pages        = {1983--1999},
  year         = {2024},
}

@article{CenJXJST24,
  author       = {Jiazhong Cen and
                  Zekun Jiang and
                  Lingxi Xie and
                  Dongsheng Jiang and
                  Wei Shen and
                  Qi Tian},
  title        = {Consensus Synergizes With Memory: {A} Simple Approach for Anomaly
                  Segmentation in Urban Scenes},
  journal      = {IEEE Transactions on Circuits and Systems for Video Technology},
  volume       = {34},
  number       = {2},
  pages        = {1086--1097},
  year         = {2024},
}

@article{zhong2019centralized,
  title={Centralized large margin cosine loss for open-set deep palmprint recognition},
  author={Zhong, Dexing and Zhu, Jinsong},
  journal={IEEE Transactions on Circuits and Systems for Video Technology},
  volume={30},
  number={6},
  pages={1559--1568},
  year={2019},
  publisher={IEEE}
}

@article{leng2019survey,
  title={A survey of open-world person re-identification},
  author={Leng, Qingming and Ye, Mang and Tian, Qi},
  journal={IEEE Transactions on Circuits and Systems for Video Technology},
  volume={30},
  number={4},
  pages={1092--1108},
  year={2019},
  publisher={IEEE}
}

@article{zhong2022bidirectional,
  title={Bidirectional spatio-temporal feature learning with multiscale evaluation for video anomaly detection},
  author={Zhong, Yuanhong and Chen, Xia and Hu, Yongting and Tang, Panliang and Ren, Fan},
  journal={IEEE Transactions on Circuits and Systems for Video Technology},
  volume={32},
  number={12},
  pages={8285--8296},
  year={2022},
  publisher={IEEE}
}

@article{sun2024classifier,
  title={Classifier-head Informed Feature Masking and Prototype-based Logit Smoothing for Out-of-Distribution Detection},
  author={Sun, Zhuohao and Qiu, Yiqiao and Tan, Zhijun and Zheng, Weishi and Wang, Ruixuan},
  journal={IEEE Transactions on Circuits and Systems for Video Technology},
  year={2024},
  publisher={IEEE}
}

@article{sun2022moep,
  title={MoEP-AE: Autoencoding mixtures of exponential power distributions for open-set recognition},
  author={Sun, Jiayin and Wang, Hong and Dong, Qiulei},
  journal={IEEE Transactions on Circuits and Systems for Video Technology},
  volume={33},
  number={1},
  pages={312--325},
  year={2022},
  publisher={IEEE}
}

@article{JiangZWH23,
  author       = {Guosong Jiang and
                  Pengfei Zhu and
                  Yu Wang and
                  Qinghua Hu},
  title        = {OpenMix+: Revisiting Data Augmentation for Open Set Recognition},
  journal      = {IEEE Transactions on Circuits and Systems for Video Technology},
  volume       = {33},
  number       = {11},
  pages        = {6777--6787},
  year         = {2023},
}

@inproceedings{huang2021mos,
  title={Mos: Towards scaling out-of-distribution detection for large semantic space},
  author={Huang, Rui and Li, Yixuan},
  booktitle={Proceedings of the IEEE/CVF Conference on Computer Vision and Pattern Recognition},
  pages={8710--8719},
  year={2021}
}

@inproceedings{
djurisic2023extremely,
title={Extremely Simple Activation Shaping for Out-of-Distribution Detection},
author={Andrija Djurisic and Nebojsa Bozanic and Arjun Ashok and Rosanne Liu},
booktitle={The Eleventh International Conference on Learning Representations },
year={2023},
url={https://openreview.net/forum?id=ndYXTEL6cZz}
}

@inproceedings{yu2023block,
  title={Block selection method for using feature norm in out-of-distribution detection},
  author={Yu, Yeonguk and Shin, Sungho and Lee, Seongju and Jun, Changhyun and Lee, Kyoobin},
  booktitle={Proceedings of the IEEE/CVF Conference on Computer Vision and Pattern Recognition},
  pages={15701--15711},
  year={2023}
}

@article{tack2020csi,
  title={Csi: Novelty detection via contrastive learning on distributionally shifted instances},
  author={Tack, Jihoon and Mo, Sangwoo and Jeong, Jongheon and Shin, Jinwoo},
  journal={Advances in neural information processing systems},
  volume={33},
  pages={11839--11852},
  year={2020}
}

@inproceedings{wei2022mitigating,
  title={Mitigating neural network overconfidence with logit normalization},
  author={Wei, Hongxin and Xie, Renchunzi and Cheng, Hao and Feng, Lei and An, Bo and Li, Yixuan},
  booktitle={International conference on machine learning},
  pages={23631--23644},
  year={2022},
  organization={PMLR}
}

@inproceedings{wang2022vim,
  title={Vim: Out-of-distribution with virtual-logit matching},
  author={Wang, Haoqi and Li, Zhizhong and Feng, Litong and Zhang, Wayne},
  booktitle={Proceedings of the IEEE/CVF conference on computer vision and pattern recognition},
  pages={4921--4930},
  year={2022}
}

@article{sun2021react,
  title={React: Out-of-distribution detection with rectified activations},
  author={Sun, Yiyou and Guo, Chuan and Li, Yixuan},
  journal={Advances in Neural Information Processing Systems},
  volume={34},
  pages={144--157},
  year={2021}
}

@inproceedings{sun2022dice,
  title={Dice: Leveraging sparsification for out-of-distribution detection},
  author={Sun, Yiyou and Li, Yixuan},
  booktitle={European Conference on Computer Vision},
  pages={691--708},
  year={2022},
  organization={Springer}
}

@inproceedings{ahn2023line,
  title={Line: Out-of-distribution detection by leveraging important neurons},
  author={Ahn, Yong Hyun and Park, Gyeong-Moon and Kim, Seong Tae},
  booktitle={2023 IEEE/CVF Conference on Computer Vision and Pattern Recognition (CVPR)},
  pages={19852--19862},
  year={2023},
  organization={IEEE}
}

@article{zhu2022boosting,
  title={Boosting out-of-distribution detection with typical features},
  author={Zhu, Yao and Chen, YueFeng and Xie, Chuanlong and Li, Xiaodan and Zhang, Rong and Xue, Hui and Tian, Xiang and Chen, Yaowu and others},
  journal={Advances in Neural Information Processing Systems},
  volume={35},
  pages={20758--20769},
  year={2022}
}

@article{zheng2023out,
  title={Out-of-distribution detection learning with unreliable out-of-distribution sources},
  author={Zheng, Haotian and Wang, Qizhou and Fang, Zhen and Xia, Xiaobo and Liu, Feng and Liu, Tongliang and Han, Bo},
  journal={Advances in Neural Information Processing Systems},
  volume={36},
  pages={72110--72123},
  year={2023}
}

@article{du2024dream,
  title={Dream the impossible: Outlier imagination with diffusion models},
  author={Du, Xuefeng and Sun, Yiyou and Zhu, Jerry and Li, Yixuan},
  journal={Advances in Neural Information Processing Systems},
  volume={36},
  year={2024}
}

@inproceedings{
wu2023energybased,
title={Energy-based Out-of-Distribution Detection for Graph Neural Networks},
author={Qitian Wu and Yiting Chen and Chenxiao Yang and Junchi Yan},
booktitle={The Eleventh International Conference on Learning Representations },
year={2023},
url={https://openreview.net/forum?id=zoz7Ze4STUL}
}

@inproceedings{macqueen1967some,
  title={Some methods for classification and analysis of multivariate observations},
  author={MacQueen, James and others},
  booktitle={Proceedings of the fifth Berkeley symposium on mathematical statistics and probability},
  volume={1},
  number={14},
  pages={281--297},
  year={1967},
  organization={Oakland, CA, USA}
}

@article{cuturi2013sinkhorn,
  title={Sinkhorn distances: Lightspeed computation of optimal transport},
  author={Cuturi, Marco},
  journal={Advances in neural information processing systems},
  volume={26},
  year={2013}
}

@InProceedings{Graham_2023_CVPR,
    author    = {Graham, Mark S. and Pinaya, Walter H.L. and Tudosiu, Petru-Daniel and Nachev, Parashkev and Ourselin, Sebastien and Cardoso, Jorge},
    title     = {Denoising Diffusion Models for Out-of-Distribution Detection},
    booktitle = {Proceedings of the IEEE/CVF Conference on Computer Vision and Pattern Recognition (CVPR) Workshops},
    month     = {June},
    year      = {2023},
    pages     = {2947-2956}
}

@article{ho2020denoising,
  title={Denoising diffusion probabilistic models},
  author={Ho, Jonathan and Jain, Ajay and Abbeel, Pieter},
  journal={Advances in neural information processing systems},
  volume={33},
  pages={6840--6851},
  year={2020}
}
\vfill

\end{document}